\documentclass[journal]{IEEEtran}

\ifCLASSINFOpdf
\else
\fi

\usepackage{multirow}
\usepackage{graphicx}
\graphicspath{{./}{./figures/}{./rev_figures/}}
\usepackage{fdsymbol}
\usepackage{amsmath,mathtools}
\usepackage{array}
\usepackage{xspace}
\usepackage{xcolor}
\usepackage{tablefootnote}
\usepackage{soul}
\usepackage{url}
\usepackage{accents}
\usepackage[flushleft]{threeparttable}

\newcommand{\Tref}[1]{Table~\ref{#1}}

\newcommand{\Eref}[1]{Equation~\eqref{#1}}
\newcommand{\fref}[1]{Fig.~\ref{#1}}
\newcommand{\Fref}[1]{Figure~\ref{#1}}

\usepackage{xcolor}
\newcounter{todos}
\AtEndDocument{\ifnum\value{todos}>0 \PackageWarning{TODOS}{There are \arabic{todos} todos left in this paper! Fix them before submitting the paper!} \fi}

\newcommand{\argmin}{\mathop{\mathrm{argmin}}\limits}

\begin{document}
\bstctlcite{IEEEexample:BSTcontrol}
\title{Fixed Viewpoint Mirror Surface Reconstruction under an Uncalibrated Camera}

\author{Kai Han, 
        Miaomiao Liu,
        Dirk Schnieders,
        and Kwan-Yee K. Wong
\thanks{K. Han is with the University of Bristol, Bristol, United Kingdom (e-mail: kai.han@bristol.ac.uk).}%
\thanks{M. Liu is with The Australian National University, Canberra, Australia (e-mail: miaomiao.liu@anu.edu.au).}%
\thanks{D. Schnieders and K.-Y. K. Wong are with The University of Hong Kong, Hong Kong, China. (e-mail: sdirk@cs.hku.hk; kykwong@cs.hku.hk).}
}

\markboth{IEEE Transactions on Image Processing}%
{Han \MakeLowercase{\textit{et al.}}: Fixed Viewpoint Mirror Surface Reconstruction under an Uncalibrated Camera}

\maketitle

\begin{abstract}
This paper addresses the problem of mirror surface reconstruction, and  proposes a solution based on observing the reflections of a moving reference plane on the mirror surface. Unlike previous approaches which require tedious calibration, our method can recover the camera intrinsics, the poses of the reference plane, as well as the mirror surface from the observed reflections of the reference plane under at least three unknown distinct poses. We first show that the 3D poses of the reference plane can be estimated from the reflection correspondences established between the images and the reference plane. We then form a bunch of 3D lines from the reflection correspondences, and derive an analytical solution to recover the \emph{line projection matrix}. We transform the line projection matrix to its equivalent camera projection matrix, and propose a cross-ratio based formulation to optimize the camera projection matrix by minimizing reprojection errors. The mirror surface is then reconstructed based on the optimized cross-ratio constraint. Experimental results on both synthetic and real data are presented, which demonstrate the feasibility and accuracy of our method. 
\end{abstract}

\begin{IEEEkeywords}
Mirror surface, reconstruction, reflection, light path.
\end{IEEEkeywords}

\IEEEpeerreviewmaketitle

\section{Introduction}
\IEEEPARstart{3}{D} reconstruction of diffuse surfaces has enjoyed tremendous success. These surfaces reflect light from a single incident ray to many rays in all directions, resulting in a constant appearance regardless of the observer's viewpoint. Methods for diffuse surface reconstruction can therefore rely on the appearance of the object.

This paper considers mirror surfaces, which exhibit \emph{specular} reflections and whose appearances are a reflection of the surrounding environment. Under (perfect) specular reflection, an incoming ray is reflected to a single outgoing ray. This special characteristic makes the appearance of a mirror surface viewpoint dependent, and renders diffuse surface reconstruction methods useless. Meanwhile, there exist many objects with a mirror surface in the man-made environment. The study of mirror surface reconstruction is therefore an important problem in computer vision.

\begin{figure}[t]
\tabcolsep=0.01\linewidth
\centering 
\begin{tabular}{ >{\centering\arraybackslash} m{0.48\linewidth} >{\centering\arraybackslash} m{0.48\linewidth}}
  \includegraphics[width=1\linewidth]{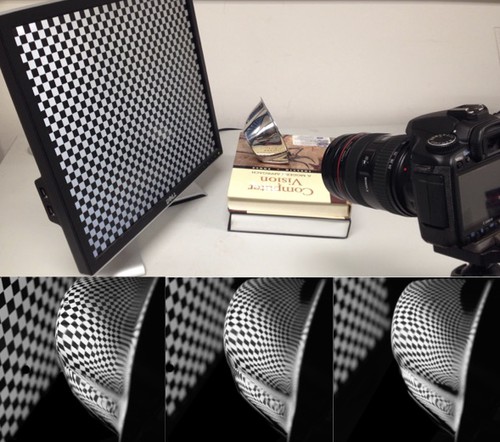} &
  \includegraphics[width=1\linewidth]{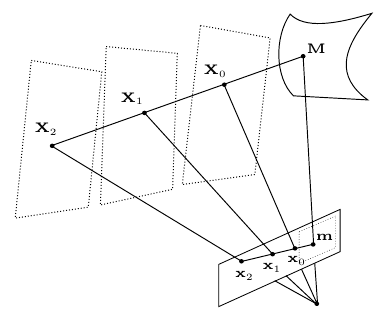}\\
(a) &
(b)\\
\end{tabular}
\caption{Real mirror surface reconstruction setup. \textmd{(a) A stationary uncalibrated camera observing the reflections of a reference plane placed at three distinct locations.
(b) Surface points can be recovered using the cross-ratio between a surface point ${\bf M}$ and its reflection correspondences $\{{\bf X}_0, {\bf X}_1, {\bf X}_2\}$ on the reference plane at three different locations.}} 
\label{fig:realsetup}
\end{figure}

In this paper, we assume the mirror surface reflects a light ray only once, and tackle the mirror surface reconstruction problem by adopting a common approach of introducing motion to the environment. Unlike previous methods which require a fully calibrated camera and known motion, we propose a novel solution based on observing the reflections of a reference plane undergoing an unknown motion with a stationary uncalibrated camera (see Fig.~\ref{fig:realsetup}(a)). Note that the checkerboard pattern in Fig.~\ref{fig:realsetup}(a) is used here only to demonstrate the overall reflection effects of the mirror surface. In our experiment, we display vertical and horizontal sweeping lines (e.g., the bottom row in Fig.~\ref{fig:realobjects}) to establish reflection correspondences between images and the reference plane.

We first show that the relative poses of the reference plane can be estimated from the reflection correspondences established between the images and the reference plane under three unknown distinct poses. This enables us to construct a 3D ray piercing the reference plane at specific positions under different poses for each image point on the mirror surface. Given the set of 3D rays and their corresponding image points, we derive an analytical solution to recover the camera projection matrix through estimating the \emph{line projection matrix}, which can be easily transformed back into a corresponding camera projection matrix. To make our solution more robust to noise, we use this closed-form solution as an initialization and optimize the camera projection matrix by minimizing reprojection errors computed based on a cross-ratio formulation for the mirror surface (see Fig.~\ref{fig:realsetup}(b)). The mirror surface is then reconstructed based on the optimized cross-ratio constraint. The key contributions of this work are
\begin{enumerate}
  \item[1.] To the best of our knowledge, the first mirror surface reconstruction solution under an unknown motion of a reference plane and an uncalibrated camera.
  \item[2.] A closed-form solution for recovering the relative 3D poses of the reference plane from reflection correspondences.
  \item[3.] A closed-form solution for estimating the camera projection matrix from  reflection correspondences.
  \item[4.] A cross-ratio based nonlinear formulation that allows a robust estimation of the camera projection matrix together with the mirror surface.
\end{enumerate}

We have presented preliminary results of this work in \cite{han2016cvpr}. This paper extends \cite{han2016cvpr} as follows: 1) We include more experimental results. In particular, we reconstructed a mirror hood and the results, in terms of quality, are similar to those of the previous experiments. This further validates the effectiveness of our approach. 2) We include details of the reference plane pose estimation from reflection correspondences and the related experimental results to make our work more self-contained. 3) We quantify the surface  of an object that can be reconstructed using our method by deriving a relation among the surface normal, the distance between the reference plane and the surface, and the size of the reference plane. 4) We include discussions on the degeneracy as well as the limitations of our methods. 5) We describe the conversion between the point projection matrix and the line projection matrix in detail. 

The rest of the paper is organized as follows. Section~\ref{relatedwork} briefly reviews existing techniques in the literature for shape recovery of mirror surfaces. Section~\ref{sec:AcquisitionSetup} describes our data acquisition setup and our closed-form solution for 3D poses estimation of the reference plane. Section~\ref{closeformsolution} introduces our closed-form solution for camera projection matrix estimation. Section~\ref{robustestimation} describes our cross-ratio based nonlinear formulation. 
Section~\ref{discussion} quantifies the surface that can be reconstructed by our method and discusses the degeneracy.
Experimental results are presented in Section~\ref{evaluation}, followed by conclusions in Section~\ref{conclusion}.

\section{Related Work}
\label{relatedwork}
Great efforts have been devoted to the problem of mirror surface recovery \cite{Balzer2010measurement, ihrke2010cgf, reshetouski2013}. Based on the assumed prior knowledge, shape recovery methods for mirror surfaces can be classified into those assuming an {\em unknown distant} ~environment and those assuming a {\em known nearby} environment.

Under an {\em unknown distant} environment, a set of methods referred to as shape from specular flow (SFSF) have been proposed. In \cite{oren1996ijcv}, Oren and Nayar successfully recovered a 3D curve on the object surface by tracking the trajectory of the reflection of a light source on the mirror surface. However, it is difficult to track a complete trajectory since the reflected feature will be greatly distorted near the occluding boundary of an object. Roth and Black \cite{roth2006cvpr} introduced the concept of specular flow and derived its relation with the 3D shape of a mirror surface. Although they only recovered a surface with a parametric representation (e.g., sphere), their work provided a theoretical basis for the later methods. In \cite{adato2007iccv, adato2010pami}, Adato \emph{et al.} showed that under far-field illumination and large object-environment distance, the observed specular flow can be related to surface shape through a pair of coupled nonlinear partial differential equations (PDEs). Vasilyev \emph{et al.} \cite{Vasilyev2011cvpr} further suggested that it is possible to reconstruct a smooth surface from one specular flow by inducing integrability constraints on the surface normal field. In \cite{canas2009iccv}, Canas \emph{et al.} reparameterized the nonlinear PDEs as linear equations and derived a more manageable solution. Although SFSF achieves a theoretical breakthrough in shape recovery of mirror surfaces, the issues in tracking dense specular flow and solving PDEs still hinder their practical use. In \cite{sankaranarayanan2010cvpr}, Sankaranarayanan \emph{et al.} developed an approach that uses sparse specular reflection correspondences instead of specular flow to recover a mirror surface linearly. Their proposed method is more practical than the traditional SFSF methods. Nevertheless, their method requires quite a number of specular reflection correspondences across different views, which are difficult to obtain due to the distorted reflections on the mirror surface.

Under a {\em known nearby} environment, a different set of methods for shape recovery of mirror surfaces can be derived. The majority of these methods are based on the smoothness assumption on the mirror surface. Under this assumption, one popular way is to formulate the surface recovery into the problem of solving PDEs. In \cite{Savarese2001cvpr, Savarese2002eccv}, Savarese and Perona demonstrated that local surface geometry of a mirror surface can be determined by analyzing the local differential properties of the reflections of two calibrated lines. Following the same fashion, Rozenfeld \emph{et al.} \cite{Rozenfeld2011pami} explored the 1D homography relation between the calibrated lines and the reflections using sparse correspondences. Depth and first order local shape are estimated by minimizing a statistically correct measure, and a dense 3D surface is then constructed by performing a constrained interpolation. In \cite{Liu2015pami}, Liu \emph{et al.} proved that a smooth mirror surface can be determined up to a two-fold ambiguity from just one reflection view of a calibrated reference plane. Another way to formulate the mirror surface recovery is by employing normal consistency property to refine visual hull and/or integrate normal field. In \cite{bonfort2003iccv}, Bonfort and Sturm introduced a voxel carving method to reconstruct a mirror surface using a normal consistency criterion derived from the reflections of some calibrated reference planes. In order to get a better view for shape recovery, they further proposed that the camera does not need to face the reference plane, and the shape can be well recovered by using a mirror to calibrate the poses of the reference plane \cite{bonfort2006accv, Sturm2006accv}. In \cite{Nehab2008cvpr}, Nehab \emph{et al.} formulated the shape recovery as an image matching problem by minimizing a cost function based on normal consistency. In \cite{Weinmann2013iccv}, Weinmann \emph{et al.} employed a turntable setup with multiple cameras and displays, which enables the calculation of the normal field for each reflection view. The 3D surface is then estimated by a robust multi-view normal field integration technique. In \cite{Balzeretal2014}, Balzer \emph{et al}. deployed a room-sized cube consisting of six walls that encode/decode specular correspondences based on a phase shift method. The surface is then recovered by integration of normal fields. Tin \emph{et al.} \cite{Tin2016iccp} introduced a two-layer LCD setup, which contains a pair of perpendicular linear polarizers for establishing correspondences between the illumination rays and camera rays. After calibrating the camera and LCDs, the surface can be reconstructed by solving a joint optimization problem. In~\cite{lu2019iccp}, Lu \emph{et al}. introduced a setup to generate a polarization field using a commercial LCD with the top polarizer removed and modeled the liquid crystals as polarization rotators. Another approach is to reconstruct the individual light paths based on the law of reflection. Kutulakos and Steger \cite{Kutulakos2008ijcv} showed that a point on a mirror surface can be recovered if the positions of two reference points are known in space and reflected to the same image point in a single view, or the positions of two reference points are known and are reflected by the same surface point to two different views. 

Note that calibration plays an important role in all the above methods that assume a known nearby environment. Commonly, a reference plane with a known pattern is used as the known environment. In order to produce a good view of its reflections on the specular surface, the reference plane is often placed side-by-side with the camera. This results in the camera not being able to see the reference plane directly, making the calibration of the setup non-trivial. Traditional methods calibrate the poses of the reference plane by introducing an extra reference plane in the field of view of the camera, and an extra camera looking at both reference planes. In \cite{Sturm2006accv}, Sturm and Bonfort used a planar mirror to allow the camera to see the reference plane through reflection. The pose of the reference plane can be obtained by placing the auxiliary mirror in at least three different positions. Generally, multiple reference plane positions are needed for recovering a large area of the mirror surface. However, the literature becomes comparatively sparse when it comes to automatic pose estimation of the reference plane in mirror surface recovery. Liu \emph{et al.} \cite{Liu2010accv}  proposed an automatic motion estimation method by constraining the motion of the reference plane to a pure translation. Although they can achieve a simple closed-form solution for the motion estimation problem, their method cannot handle general motion. Besides, their method requires calibrating the intrinsics of the camera as well as the initial pose of the reference plane. In fact, most, if not all, of the methods that assume a known nearby environment require the camera(s) to be fully calibrated. In contrast, we neither require the calibration of the reference plane poses, nor require the calibration of the camera, and make no assumption on the smoothness of the mirror surface. Our proposed approach can automatically calibrate the setup as well as reconstruct the mirror surface using the observed reflections of the reference plane.

A cross-ratio constraint has been used to estimate mirror position and camera pose for axial non-central catadioptric systems \cite{Perdigoto2013cviu,ramalingam2005cvpr}, and produce more point correspondences in the context of 3D reconstruction \cite{Ramalingam_2015_CVPR}. In this work, we incorporate cross-ratio constraint in our formulation to simultaneously optimize the camera projection matrix and recover the mirror surface.

\section{Acquisition Setup and Plane Pose Estimation} \label{sec:AcquisitionSetup}

\subsection{Acquisition Setup}
\label{sec:synsetup}

\begin{figure} [htbp]
   \begin{center}
  \includegraphics[width=0.8\linewidth]{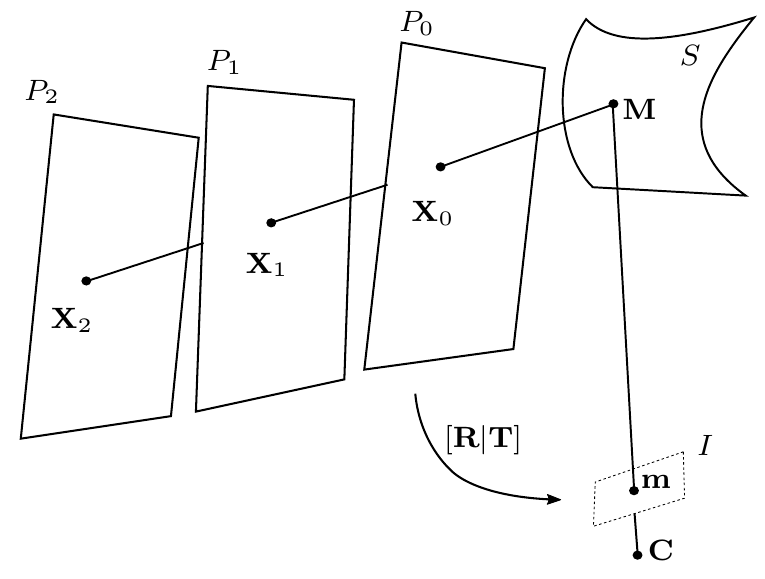}
   \end{center}
   \caption{Setup used for mirror surface reconstruction. A pinhole camera centered at ${\bf C}$ observes the mirror surface $S$ which reflects a moving reference plan placed at three different locations $P_0$, $P_1$ and $P_2$. \textmd{Refer to Section~\ref{sec:synsetup} for notations and definitions.}}
   \label{fig:syn_setup}
\end{figure}

\Fref{fig:syn_setup} shows the setup used for mirror surface reconstruction. Consider a pinhole camera centered at ${\bf C}$ observing the reflections of a moving reference plane on a mirror surface $S$. Let ${\bf X}_0$ be a point on the plane at its initial pose, denoted by $P_0$, which is reflected by a point ${\bf M}$ on $S$ to a point ${\bf m}$ on the image plane $I$. Suppose the reference plane undergoes an unknown rigid body motion, and let $P_1$ and $P_2$ denote the plane at its two new poses. Let ${\bf X}_1$ and ${\bf X}_2$ be points on $P_1$ and $P_2$, respectively, which are both reflected by ${\bf M}$ on $S$ to the same image point ${\bf m}$ on $I$. ${\bf X}_0$, ${\bf X}_1$ and ${\bf X}_2$ are referred to as \emph{reflection correspondences} of the image point ${\bf m}$.
Since reflection correspondences must lie on the same incident ray, it follows that they must be colinear in 3D space. This property will be used to derive a constraint for computing the poses of the moving reference plane relative to its initial pose (see Section~\ref{sec:planepose}). 

\subsection{Plane Pose Estimation}
\label{sec:planepose}

Referring to the setup shown in Fig.~\ref{fig:syn_setup}. Let the rigid body motion between $P_0$ and $P_1$ be denoted by (${\bf R}^1,{\bf T}^1$), where ${\bf R}^1$ and ${\bf T}^1$ are the rotation (matrix) and translation (vector) respectively. Similarly, let (${\bf R}^2,{\bf T}^2$) denote the rigid body motion between $P_0$ and $P_2$. Let $X_0$, $X_1$, and $X_2$ be points on $P_0$, $P_1$ and $P_2$, respectively, that lie on the same incident light path. The 2D coordinates of $X_i$ on the plane are ${\bf
X}^{\rm p}_i$= $[x^{\rm p}_i \ y^{\rm p}_i \ 0]^{\rm T}$, where $i\in
\{0,1,2\}$. Their 3D coordinates,
${\bf X}_i = [x_i \ y_i \ z_i]^\mathrm{T}$, $i\in \{0,1,2\}$, w.r.t $P_0$
can be written as
\begin{equation*}
\begin{split}
{\bf X}_0 &= {\bf X}^{\rm p}_0=\left[x_0 \ y_0 \ z_0\right]^\mathrm{T}=\left[x_0 \ y_0 \ 0\right]^\mathrm{T},\\
{\bf X}_1 &={\bf R}^1{\bf X}^{\rm p}_1+ {\bf T}^1 = \mathcal{M}{\bf \bar X}^{\rm p}_1,\\
{\bf X}_2 &= {\bf R}^2{\bf X}^{\rm p}_2+{\bf T}^2 =\mathcal{N}{\bf \bar X}^{\rm p}_2,
\end{split}
\end{equation*}
where
 $\mathcal{M} = \left[{\bf R}^1_{*1} \ {\bf R}^1_{*2} \ {\bf T}^1\right]$, 
$\mathcal{N} = \left[{\bf R}^2_{*1} \ {\bf R}^2_{*2} \ {\bf T}^2\right]$, 
${\bf \bar{X}}^{\rm p}_i =[x^{\rm p}_i \ y^{\rm p}_i \ 1]^\mathrm{T}$, 
and ${\bf R}^i_{*j}$ denotes the $j$th column of
${\bf R}^i, i\in \{1,2\}, j\in \{1,2\}$. 
The unknown motion parameters are now embedded in $\mathcal{M}$ and $\mathcal{N}$. Since ${\bf X}_0, \ {\bf X}_1$ and ${\bf X}_2$ are
colinear, it follows that
\begin{equation}
\begin{split}
\frac{x_1-x_0}{x_2-x_0} &= \frac{y_1-y_0}{y_2-y_0}=\frac{z_1-z_0}{z_2-z_0},\\
\frac{\mathcal{M}_{1*}{\bf \bar{X}}^{\rm p}_1 - x_0}{\mathcal{N}_{1*}{\bf \bar{X}}^\mathrm{p}_2-x_0} &=\frac{\mathcal{M}_{2*}{\bf \bar{X}}^\mathrm{p}_1-y_0}{\mathcal{N}_{2*}{\bf \bar{X}}^\mathrm{p}_2-y_0}=\frac{\mathcal{M}_{3*}{\bf \bar{X}}^\mathrm{p}_1}{\mathcal{N}_{3*}{\bf
\bar{X}}^\mathrm{p}_2},\label{eq:colinear}
\end{split}
\end{equation} where $\mathcal{M}_{i*}$ and $\mathcal{N}_{i*}$ denote the $i$th row of $\mathcal{M}$ and $\mathcal{N}$ respectively. The following two constraints can be derived from \Eref{eq:colinear}:
\begin{equation}
\begin{cases}
({\bf {\bar X}}^\mathrm{p}_2)^\mathrm{T}\mathcal{A}{\bf \bar{X}}^\mathrm{p}_1-x_0({\bf \bar{X}}^\mathrm{p}_2)^\mathrm{T}\mathcal{N}_{3*}^\mathrm{T}+x_0({\bf {\bar X}}^\mathrm{p}_1)^\mathrm{T} \mathcal{M}_{3*}^\mathrm{T}=0,\\
({\bf {\bar X}}^\mathrm{p}_2)^\mathrm{T}\mathcal{B}{\bf
\bar{X}}^\mathrm{p}_1-y_0({\bf \bar{X}}^\mathrm{p}_2)^\mathrm{T}\mathcal{N}_{3*}^\mathrm{T}+y_0({\bf \bar{X}}^\mathrm{p}_1)^\mathrm{T}\mathcal{M}_{3*}^\mathrm{T}=0,
\end{cases}\label{eq:cons2}
\end{equation}
where $\mathcal{A}=
\mathcal{N}_{3*}^\mathrm{T}\mathcal{M}_{1*} - \mathcal{N}_{1*}^\mathrm{T}\mathcal{M}_{3*}$ and $\mathcal{B}=\mathcal{N}_{3*}^\mathrm{T}\mathcal{M}_{2*} -\mathcal{N}_{2*}^\mathrm{T}\mathcal{M}_{3*}$.

Given $3 \times m$ points ${\bf \bar{X}}_{ij}^\mathrm{p}=[x_{ij}^\mathrm{p} \ y_{ij}^\mathrm{p} \ z_i]^\mathrm{T}$, where $0\leq i\leq 2$, $1\leq j\leq m$, $z_0 = 0$ and $z_{i\in\{1, 2\}}=1$, we can formulate the problem as solving a linear system
\begin{equation}
{\bf EW}=\bf{0},
\end{equation}
where 
\begin{equation*}
\makeatletter\def\f@size{7}\check@mathfonts
\def\maketag@@@#1{\hbox{\m@th\large\normalfont#1}}
{\bf E} =\left[\begin{smallmatrix}
({\bf \bar{X}}^\mathrm{p}_{21})^\mathrm{T}\otimes({\bf \bar{X}}^\mathrm{p}_{11})^\mathrm{T}& {\bf 0}^\mathrm{T} & -x^\mathrm{p}_{01}({\bf \bar{X}}^\mathrm{p}_{21})^\mathrm{T}  & -x^\mathrm{p}_{01}({\bf \bar{X}}^\mathrm{p}_{11})^\mathrm{T} \\
         {\bf 0}^\mathrm{T} & ({\bf \bar{X}}^\mathrm{p}_{21})^\mathrm{T}\otimes({\bf \bar{X}}^\mathrm{p}_{11})^\mathrm{T}& -y^\mathrm{p}_{01}({\bf \bar{X}}^\mathrm{p}_{21})^\mathrm{T}  & -y^\mathrm{p}_{01}({\bf \bar{X}}^\mathrm{p}_{11})^\mathrm{T} \\
\vdots& \vdots & \vdots & \vdots \\
({\bf \bar{X}}^\mathrm{p}_{2m})^\mathrm{T}\otimes({\bf \bar{X}}^\mathrm{p}_{1m})^\mathrm{T}& {\bf 0}^\mathrm{T} & -x^\mathrm{p}_{0m}({\bf \bar{X}}^\mathrm{p}_{2m})^\mathrm{T}  & -x^\mathrm{p}_{0m}({\bf \bar{X}}^\mathrm{p}_{1m})^\mathrm{T} \\
         {\bf 0}^\mathrm{T} & ({\bf \bar{X}}^{\rm p}_{2m})^\mathrm{T}\otimes({\bf \bar{X}}^{\rm p}_{1m})^\mathrm{T}& -y^\mathrm{p}_{0m}({\bf \bar{X}}^\mathrm{p}_{2m})^\mathrm{T}  & -y^\mathrm{p}_{0m}({\bf \bar{X}}^\mathrm{p}_{1m})^\mathrm{T} \\
 \end{smallmatrix}\right],
\end{equation*}
\begin{equation}
{\bf W} =\left[
\begin{smallmatrix}
\mathcal{A}_{1*}& \mathcal{A}_{2*} & \mathcal{A}_{3*} &
\mathcal{B}_{1*} &\mathcal{B}_{2*} & \mathcal{B}_{3*}&
\mathcal{N}_{3*} &\mathcal{M}_{3*}
\end{smallmatrix}
\right]^\mathrm{T},\label{eq:defineW}
\end{equation}
$\mathcal{A}_{i*}$ and $\mathcal{B}_{i*}$ denote the $i$th row of  $\mathcal{A}$ and $\mathcal{B}$ respectively, and $\otimes$ denotes Kronecker tensor product. ${\bf W}$ contains $24$ unknowns in total. Since each incident ray provides two constraints, we need at least 12 incident rays (i.e., $3 \times 12$ reflection correspondences) to solve all the unknowns. 

Note that the $21$st and $24$th columns of ${\bf E}$ are identical, the nullity of {\bf E} must be  two in order to have a non-trivial solution. Therefore, we first apply SVD to get a solution space
spanned by two solution basis vectors, ${\bf d}_1$ and ${\bf d}_2$. 
We then parameterize $\bf W$ as
\begin{equation}
 {\bf W} = \alpha({\bf d}_1+\beta{\bf d}_2),\label{eq:w_vectorBases}
\end{equation}
where $\alpha$ and $\beta$ are two scale parameters. Now there are $26$ unknowns in total. By enforcing the element-wise equality of~\Eref{eq:defineW} and~\Eref{eq:w_vectorBases}, we have $18$ bilinear and $6$ linear equations to solve $\mathcal{M}$, $\mathcal{N}$, $\alpha$ and $\beta$. Furthermore, we have $rank(\mathcal{N}_{3*}^\mathrm{T}\mathcal{M}_{1*})\leq1$ and $rank( -\mathcal{N}_{1*}^\mathrm{T}\mathcal{M}_{3*})\leq1$ since $rank(\mathcal{N}_{3*}^\mathrm{T}\mathcal{M}_{1*})\leq min(rank(\mathcal{N}_{3*}^\mathrm{T}),rank(\mathcal{M}_{1*}^\mathrm{T}))=1$. Thus,
\begin{equation}
\begin{aligned}
rank(\mathcal{A}) &= rank(\mathcal{N}_{3*}^\mathrm{T}\mathcal{M}_{1*}-\mathcal{N}_{1*}^\mathrm{T}\mathcal{M}_{3*})\nonumber \\
&\leq rank(\mathcal{N}_{3*}^\mathrm{T}\mathcal{M}_{1*})+ rank(
-\mathcal{N}_{1*}^\mathrm{T}\mathcal{M}_{3*})\nonumber \\
&\leq 2.
\end{aligned}
\end{equation}
Similarly, we can show $rank(\mathcal{B})\leq2$. Obviously, not all of the obtained constraints are independent, and new constraints should be applied in order to solve all the unknowns. Since the first two columns of $\mathcal{M}$ and $\mathcal{N}$ come from the first two columns of ${\bf R}^1$ and ${\bf R}^2$ respectively, the orthonomality property will provide $6$ extra constraints, which leads to a closed-form solution for the unknown motion parameters and the two scale parameters. We use the \emph{Symbolic Math Toolbox in Matlab} to solve them (refer to the supplementary for more details).

\section{Projection Matrix Estimation}  \label{closeformsolution}
In this section, we first briefly review the line projection matrix. We then derive a linear method for obtaining a closed-form solution to the line projection matrix of a camera from the reflection correspondences.

\subsection{Line Projection Matrix}\label{sec:lineprojection}
\begin{figure}[h]
\tabcolsep=0.01\linewidth
\centering 
\begin{tabular}{ 
>{\centering\arraybackslash} m{0.46\linewidth} 
>{\centering\arraybackslash} m{0.4\linewidth}}
    \includegraphics{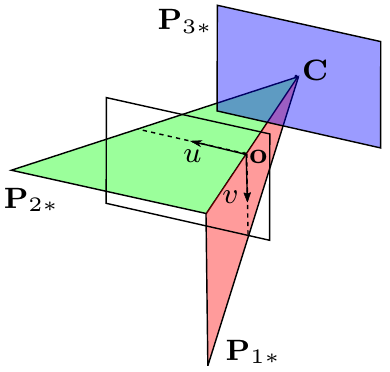}&
    \includegraphics{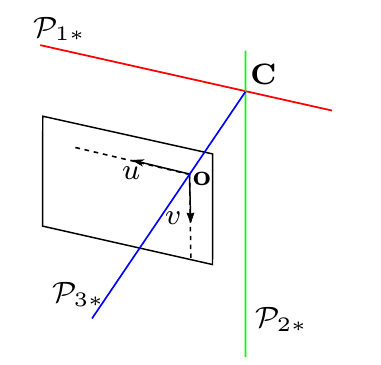}\\
(a) &
(b)\\
\end{tabular}
\caption{Visualization of the point and line projection matrices. \textmd{(a) Rows of a point projection matrix represent planes that intersect at the optical center $\bf C$ of the camera. (b) Dually, rows of a line projection matrix represent lines that intersect at the optical center.}}
\label{fig:visulizationp}
\end{figure}

Using homogeneous coordinates, a linear mapping can be defined for mapping a point ${\bf X}$ in 3D space to a point ${\bf x}$ in a 2D image, i.e.,
\begin{equation}
   {\bf x} = \bf{P}{\bf X}, %
   \label{eq:pproj}
\end{equation}
where ${\bf P}$ is a $3\times 4$ matrix known as the camera (point) projection matrix. Similarly, using Pl\"{u}cker coordinates\footnote{A brief review of Pl\"{u}cker coordinates is given in the supplementary.}, a linear mapping can be defined for mapping a line $\mathcal{L}$ in 3D space to a line ${\bf l}$ (in homogeneous coordinates) in a 2D image, i.e.,
\begin{equation}
   {\bf l} = \mathcal{P \bar L}, 
   \label{eq:lproj}
\end{equation}
where $\mathcal P$ is a $3\times 6$ matrix known as the \emph{line projection matrix} and $\mathcal{\bar L}$ is the \emph{dual Pl\"{u}cker vector}\footnote{Given a Pl\"{u}cker vector $\mathcal{L}=\left[l_{1}\ l_{2}\ l_{3}\ l_{4}\ l_{5}\ l_{6}\right]^{\mathrm{T}}$, its dual Pl\"{u}cker vector is ${\mathcal{\bar L}}=\left[l_{5}\ l_{6}\ l_{4}\ l_{3}\ l_{1}\ l_{2}\right]^{\mathrm{T}}$. } of $\mathcal{L}$. Note that each row ${\bf P}_{i*}$ ($i \in \{1, 2, 3\}$) of ${\bf P}$ represents a {\em plane} (in homogeneous coordinates) that intersects at the optical center. Dually, each row $\mathcal{P}_{i*}$ ($i \in \{1, 2, 3\}$) of 
$\mathcal{P}$ represents a {\em line} that intersects at the optical center (see Fig.~\ref{fig:visulizationp}). It follows that a valid line projection matrix must satisfy %
\begin{equation}
  \mathcal{P}_{i*} \cdot \mathcal{\bar P}_{j*} = 0 \ \ \forall \ i, j \in \{1, 2, 3\} \Leftrightarrow
  \mathcal{P} \mathcal{\bar P}^{\rm T} = {\bf 0_{\rm 3,3}},
  \label{eq:pcond}
\end{equation}
 where
$\mathcal{\bar P} = [\mathcal{\bar P}^{\rm T}_{1*} \ \mathcal{\bar P}^{\rm T}_{2*} \ \mathcal{\bar P}^{\rm T}_{3*}]^{\rm T}$.

A line projection matrix can be converted to its corresponding camera (point) projection matrix, and vice versa. Details of the conversion are given in the supplementary.

\subsection{Estimating the Line Projection Matrix} 
\label{initialization}
To estimate the line projection matrix of the camera, we first apply our method described in Section~\ref{sec:planepose} to recover the relative poses of the reference plane under three distinct poses using reflection correspondences established between the images and the reference plane. We can then form a 3D Pl\"{u}cker line $\mathcal{L}$ from the reflection correspondences of each observed point ${\bf x}$ in the image. Note that, by construction, ${\bf x}$ must lie on the projection of $\mathcal{L}$, i.e., 
\begin{equation}
    {\bf x}^{\rm T} \mathcal{P \bar L} = 0 . \label{eq:xPL=0}
\end{equation}
Given a set of 3D space lines $\{ \mathcal L_{1}, ..., \mathcal L_{n} \}$ constructed for a set of image points $\{{\bf x}_1, ..., {\bf x}_n \}$, the constraint derived in (\ref{eq:xPL=0}) can be arranged into
\begin{equation}
   \mathbf{Z} \mathbf{p}=\mathbf{0}, \label{eq:closep}
\end{equation}
where $\mathbf{p} =  [\mathcal{P}_{1*} \ \mathcal{P}_{2*} \ \mathcal{P}_{3*}]^{\rm T}$ and
\begin{equation}
   \mathbf{Z} = \left[ {\begin{array}{*{20}{c}}
         { {\bf x}_1^{\rm T} \otimes \bar {\mathcal{L}}_1^{\rm T}}\\
         \vdots \\
         { {\bf x}_n^{\rm T} \otimes \mathcal{\bar L}_n^{\rm T}}
      \end{array}} \right].
   \end{equation}
The line projection matrix of the camera can then be estimated by solving 
\begin{equation}
   \argmin_{\mathbf{p}} \| \mathbf{Z} \mathbf{p}\|^2 
  \label{eq:linearsolution}
\end{equation}
subject to $\| \mathbf{p} \| = 1$.
The line projection matrix thus obtained can be transformed into a point projection matrix and vice versa. 
Note that, however, (\ref{eq:linearsolution}) minimizes only algebraic errors and does not enforce (\ref{eq:pcond}). The solution to (\ref{eq:linearsolution}) is therefore subject to numerical instability and not robust in the presence of noise.
Instead of solving (\ref{eq:linearsolution}), we can minimize the geometric distance from each image point to the projection of the corresponding 3D line. Let ${\bf l} = [a, b, c]^{\rm T} = \mathcal{P\bar L}$ be the projection of the 3D line $\mathcal{L}$ corresponding to an image point ${\bf x} = [x_1, x_2, x_3]^{\rm T}$. $\mathcal{P}$ can be estimated by solving 
\begin{equation}
  \argmin_{\mathcal{P}} \sum\limits_{i = 1}^n {\frac{({{\bf x}_i^{\rm T} \mathcal{P}{\mathcal{\bar L }_i})^2}}{{a_i}^2 + {b_i}^2}} 
\label{eq:pldist}
\end{equation}
subject to $\mathcal{\|P\|} = 1$, where $\mathcal{\|P\|} $ is the Frobenius norm of $\mathcal{P}$. A straight-forward approach to enforce (\ref{eq:pcond}) is by incorporating it as a hard constraint in (\ref{eq:pldist}). However, experiments using a number of state-of-the-art optimization schemes show that such a solution often converges to local minima. 

\subsection{Enforcing Constraints}
\label{initialization_E}
Given a proper camera projection matrix, the corresponding line projection matrix will automatically satisfy (\ref{eq:pcond}). However, given an improper $3\times 6$ line projection matrix not satisfying (\ref{eq:pcond}), the corresponding camera projection matrix cannot be decomposed into one with proper intrinsic and extrinsic parameters. Based on this observation, we propose to enforce (\ref{eq:pcond}) through ensuring a proper decomposition of the camera projection matrix.

Consider a simplified scenario where the principal point $(u_0, v_0)$ (which is often located at the image centre) is known. After translating the image origin to the principal point, the camera projection matrix can be expressed as 
\begin{equation*}
      {\bf P} = {\bf K}[{\bf R} \ {\bf T}] =
      \begin{bmatrix}
         f_x & 0    & 0\\
         0    & f_y & 0\\
         0    & 0    & 1
      \end{bmatrix}
      \begin{bmatrix}
         r_{11} & r_{12}    & r_{13} & t_1\\
         r_{21} & r_{22}    & r_{23} & t_2\\
         r_{31} & r_{32}    & r_{33} & t_3
      \end{bmatrix},
\end{equation*}
and the corresponding line projection matrix (refer to the supplementary for the conversion) can be expressed as
\begin{equation}
      \mathcal{P} = 
      \begin{bmatrix}
         f_y & 0    & 0\\
         0    & f_x & 0\\
         0    & 0    & f_xf_y
      \end{bmatrix}\mathcal{P} ',
\end{equation}
where
\begin{equation}
      \mathcal {P}'^{\rm T}_{i*} 
      =
      \begin{bmatrix}
         \rho'_{i1}\\
         \rho'_{i2}\\
         \rho'_{i3}\\
         \rho'_{i4}\\
         \rho'_{i5}\\
         \rho'_{i6}
      \end{bmatrix}
      = (-1)^{(i+1)}
      \begin{bmatrix}
         r_{j3}t_{k} - t_{j}r_{k3}\\
         t_{j}r_{k2} - r_{j2}t_{k}\\
         r_{j2}r_{k3} - r_{j3}r_{k2}\\
         r_{j1}t_{k} - t_{j}r_{k1}\\
         r_{j1}r_{k2} - r_{j2}r_{k1}\\
         r_{j1}r_{k3} - r_{j3}r_{k1}
      \end{bmatrix},
   \end{equation}
with $i \neq j \neq k \in \{1, 2, 3\}$ and $j < k$. (\ref{eq:closep}) can then be rewritten as
\begin{equation}
   \mathbf{Z} \mathbf{p}  = \mathbf{ZD}\mathbf{p}' = \mathbf{Z}' \mathbf{p}' = 0 ,
   \label{eq:solveRT}
\end{equation}
where $\mathbf{p}' =  [\mathcal{P}'_{1*} \ \mathcal{P}'_{2*} \ \mathcal{P}'_{3*}]^{\rm T}$, ${\bf Z}' = {\bf Z D}$ and ${\bf D}$ is a $18 \times 18$ diagonal matrix with $d_{ii} = f_y$ for $i \in \{ 1,...,6 \}$, $d_{ii} = f_x$ for $i \in \{ 7,...,12 \}$, and $d_{ii}={f_xf_y}$ for $i \in \{ 13,...,18 \}$.

With known $f_x$ and $f_y$, $\mathbf{p}'$ can be estimated by solving (\ref{eq:solveRT}). Since $\mathcal{P} '$ only depends on the elements of $\bf R$ and $\bf T$, it can be converted to a point projection matrix (refer to the supplementary for the conversion) in the form of $\lambda [{\bf R}\ {\bf T}]$. The magnitude of $\lambda$ is determined by the orthogonality of $\bf R$, and its sign is determined by the sign of $t_3$. Hence, given the camera intrinsics, the camera extrinsics can be recovered using the reflection correspondences. 

In Section~\ref{robustestimation}, we tackle the problem of unknown camera intrinsics by formulating the problem into a nonlinear optimization by minimizing reprojection errors computed based on a cross-ratio formulation for the mirror surface. For initialization purpose, we assume $(u_0, v_0)$ being located at the image center, and $f_x = f_y = f$. We choose a rough range of $f$ and for each sample value of $f$ within the range, we estimate $\bf R$ and $\bf T$ by solving (\ref{eq:solveRT}). The point to line distance criterion in (\ref{eq:pldist}) is applied to find the best focal length $f'$. A camera projection matrix can then be constructed using $f'$, $(u_0, v_0)$, ${\bf R}$ and ${\bf T}$ that satisfies all the above mentioned constraints.

\section{Cross-ratio Based Formulation}
\label{robustestimation}
\begin{figure}[tbp]
   \begin{center}
    \includegraphics[width=0.75\linewidth]{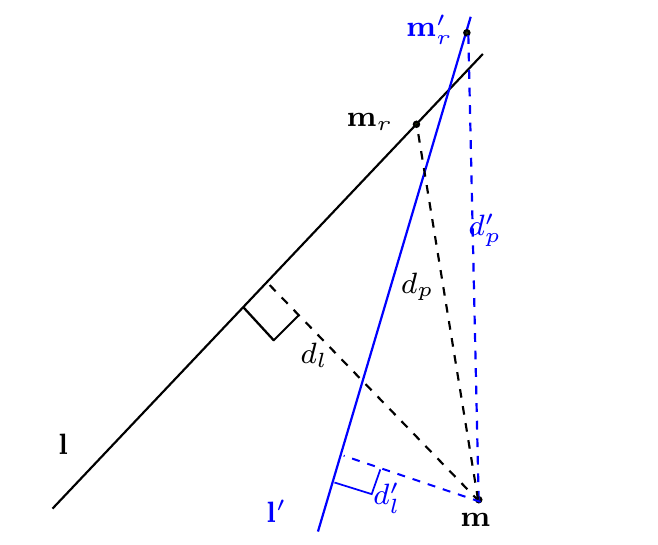}
   \end{center}
   \caption{Minimizing point-to-line distance does not guarantee minimizing point-to-point distance. \textmd{A 3D point $\bf M$ and a 3D line $\mathcal{L}$ passing through it are projected by $\mathcal{P}$ to a 2D point ${\bf m}_r$ and a 2D line ${\bf l}$, respectively. Let $\bf m$ denote the observation of $\bf M$. The distance between ${\bf m}$ and ${\bf m}_r$ is $d_p$, and the distance between ${\bf m}$ and ${\bf l}$ is $d_l$. Suppose the same 3D point $\bf M$ and 3D line $\mathcal{L}$ are projected by $\mathcal{P}'$ to ${\bf m}'_r$ and ${\bf l}'$, respectively. The distance between ${\bf m}$ and ${\bf m}'_r$ is $d'_p$, and the distance between ${\bf m}$ and ${\bf l}'$ is $d'_l$. Note that $d'_l < d_l$, but $d'_p > d_p$.}}
   \label{fig:pldist}
\end{figure}

In this section, we obtain the camera projection matrix and the mirror surface by minimizing reprojection errors. We will derive a cross-ratio based formulation for recovering a 3D point on the mirror surface from its reflection correspondences. Note that minimizing point-to-point reprojection errors can provide a stronger geometrical constraint than minimizing the point-to-line distances in (\ref{eq:pldist}) (see Fig.~\ref{fig:pldist}).

\begin{figure}[htbp]
   \begin{center}
    \includegraphics[width=0.75\linewidth]{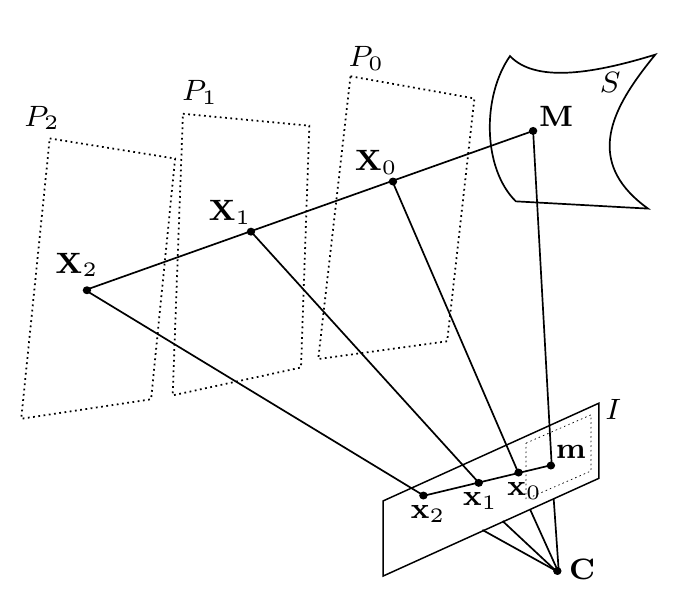}
   \end{center}
   \caption{Cross-ratio constraint. \textmd{Camera projection matrix and mirror surface points are recovered by minimizing reprojection errors computed from the cross-ratio constraint $\{{\bf M}, {\bf X}_0$; ${\bf X}_1$, ${\bf X}_2$\} = \{$\bf m$, ${\bf x}_0$; ${\bf x}_1$, ${\bf x}_2$\}, where ${\bf X}_0$, ${\bf X}_1$, ${\bf X}_2$ are the reflection correspondences of $\bf M$ on the reference plane under three different poses (i.e., $P_0$, $P_1$ and $P_2$) and $\bf m$, ${\bf x}_0$, ${\bf x}_1$, ${\bf x}_2$ are their projections on the image plane. Note that ${\bf X}_0$, ${\bf X}_1$, ${\bf X}_2$ may not be visible to the camera.}}
   \label{fig:cross_ratio}
\end{figure}

Consider a point ${\bf M}$ on the mirror surface (see Fig.~\ref{fig:cross_ratio}). Let ${\bf X}_0$, ${\bf X}_1$ and ${\bf X}_2$ be its reflection correspondences on the reference plane under three distinct poses, denoted by $P_0$, $P_1$ and $P_2$, respectively. Suppose ${\bf M}$, ${\bf X}_0$, ${\bf X}_1$ and ${\bf X}_2$ are projected to the image as ${\bf m}$, ${\bf x}_0$, ${\bf x}_1$ and ${\bf x}_2$ respectively. We observe that the cross-ratios $\{{\bf M}, {\bf X}_0; {\bf X}_1, {\bf X}_2\}$ and $\{{\bf m}, {\bf x}_0; {\bf x}_1, {\bf x}_2\}$ must be identical, i.e.,
\begin{equation}
 \frac{\|\overrightharpoon{{\bf X}_1{\bf M}}\|\|\overrightharpoon{{\bf X}_2{\bf X}_0}\|}{\|\overrightharpoon{{\bf X}_1{\bf X}_0}\|\|\overrightharpoon{{\bf X}_2{\bf M}}\|} 
   = \frac{\|\overrightharpoon{{\bf x}_1{\bf m}}\|\|\overrightharpoon{{\bf x}_2{\bf x}_0}\|}{\|\overrightharpoon{{\bf x}_1{\bf x}_0}\|\|\overrightharpoon{{\bf x}_2{\bf m}}\|} ,
   \label{eq:xratio}
\end{equation}
where $\overrightharpoon{{\bf A}{\bf B}}$ denotes the directed ray (vector) from ${\bf A}$ to ${\bf B}$ and $\|\overrightharpoon{{\bf A}{\bf B}}\|$ is the length of the vector.
Let $s$ be the distance between ${\bf X}_2$ and ${\bf M}$ (i.e., $s = \|\overrightharpoon{{\bf X}_2{\bf M}}\|$), from (\ref{eq:xratio})
\begin{equation}
\resizebox{.88\hsize}{!}{$
   {\it s} = \frac{\|\overrightharpoon{{\bf X}_2{\bf X}_1}\|\|\overrightharpoon{{\bf X}_2{\bf X}_0}\|\|\overrightharpoon{{\bf x}_1{\bf x}_0}\|\|\overrightharpoon{{\bf x}_2{\bf m}}\|}{\|\overrightharpoon{{\bf X}_2{\bf X}_0}\|\|\overrightharpoon{{\bf x}_1{\bf x}_0}\|\|\overrightharpoon{{\bf x}_2{\bf m}}\| - \|\overrightharpoon{{\bf X}_1{\bf X}_0}\|\|\overrightharpoon{{\bf x}_2{\bf x}_0}\|\|\overrightharpoon{{\bf x}_1{\bf m}}\|}$
}.
 \label{eq:s}
\end{equation}
Given the projection matrix, ${\bf x}_0$, ${\bf x}_1$, ${\bf x}_2$ and ${\bf m}$, the surface point ${\bf M}$ can be recovered as
\begin{equation}
   {\bf M} = {\bf X}_2 + {\it s}\frac{\overrightharpoon{{\bf X}_2{\bf X}_0}}{\|\overrightharpoon{{\bf X}_2{\bf X}_0}\|} .
   \label{eq:M}
\end{equation}

We optimize the projection matrix by minimizing the reprojection errors, i.e.,
\begin{equation}
   \argmin_{\theta} \sum\limits_{i = 1}^{n}({{\bf m}_i - {\bf m}^\prime_i})^2 ,
   \label{eq:non-linear}
\end{equation}
where ${\bf m}_i$ is the observation of ${\bf M}_i$, ${\bf m}^\prime_i= {\bf P}({\theta}){\bf M}_i$, and ${\theta} = [f_x\ f_y\ u_0\ v_0\ r_x\ r_y\ r_z\ t_x\ t_y\ t_z]^{\rm T}$~\footnote{We used angle-axis representation for rotation, i.e., $[r_x\ r_y\ r_z]^{\rm T} =  \tau{\bf e}$, where $\tau$ is the rotation angle and $\bf e$ is the unit rotation axis.}. To obtain $s$ in~(\ref{eq:s}), we also project the three reflection correspondences of ${\bf m}_i$ to the image using ${\bf P}({\theta})$. We initialize ${\theta}$ using the method proposed in Section~\ref{closeformsolution}, and solve the optimization problem using the Levenberg-Marquardt method. Given the estimated projection matrix, the mirror surface can be robustly reconstructed by solving~(\ref{eq:xratio})-(\ref{eq:M}).

\section{Discussion}
\label{discussion}

\subsection{Object Analysis}

\begin{figure}[htbp]
\centering 
  \includegraphics[width=0.6\linewidth]{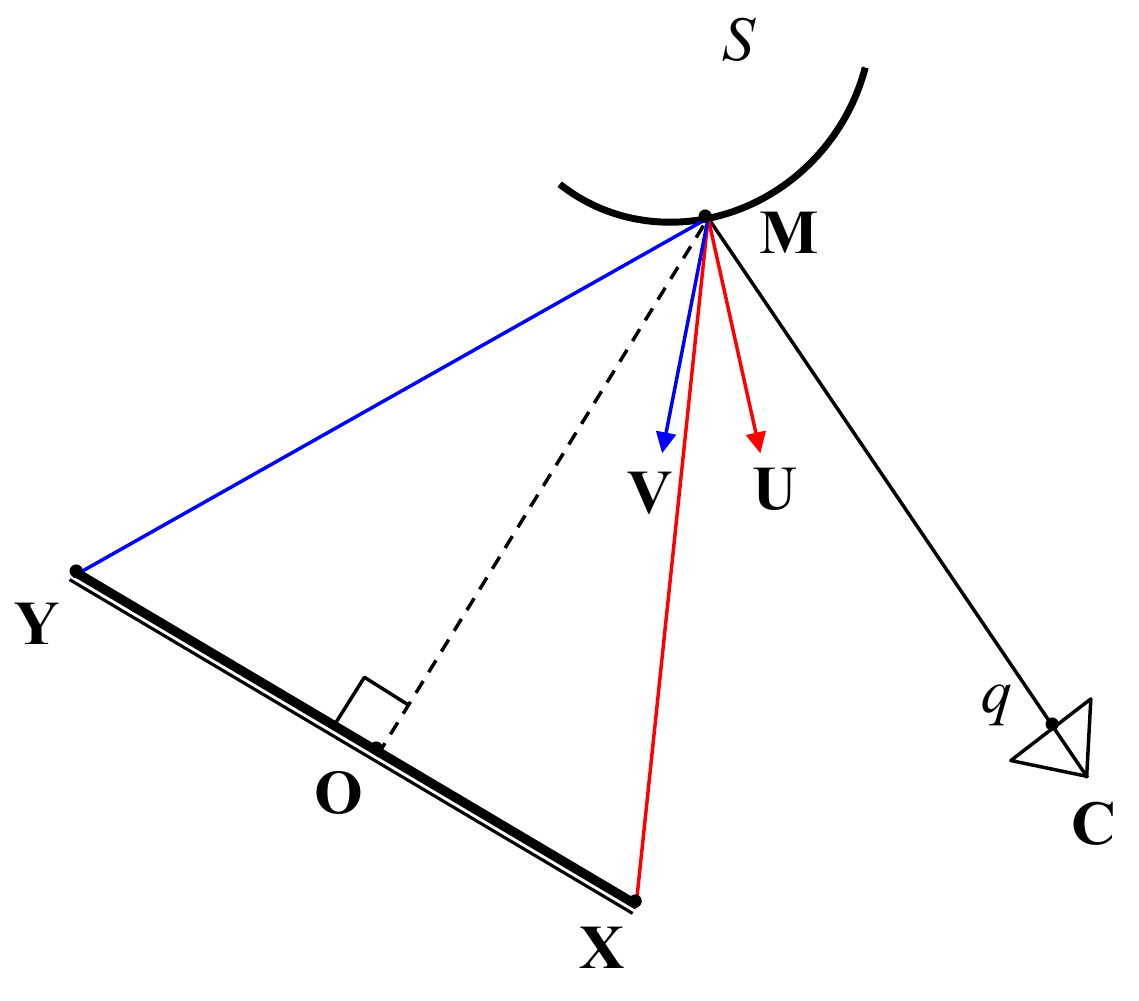}
\caption{A camera centered at $\bf C$ observes a mirror surface $S$ and a reference plane ${\bf XY}$ is placed beside the surface. \textmd{$\overrightharpoon{\bf CM}$ is the reflected ray (i.e., the visual ray of the pixel $q$) through the surface point $\bf M$. If $\overrightharpoon{\bf XM}$ is the incident ray, the normal at $\bf M$ is $\overrightharpoon{\bf MU}$, which is the bisector of $\protect\angle{\bf CMX}$. Similarly, if $\overrightharpoon{\bf YM}$ is the incident ray, the normal at $\bf M$ is $\overrightharpoon{\bf MV}$, which is the bisector of $\protect\angle{\bf CMY}$.}}
\label{fig:angle_analysis}
\end{figure}

We analyze the surface points that can be reconstructed by our proposed method using an example as shown in Fig.~\ref{fig:angle_analysis}, where a fixed camera centered at ${\bf C}$ is viewing a mirror surface $S$, and a reference plane ${\bf XY}$ is placed beside the surface. Consider the visual ray $\overrightharpoon{\bf CM}$ of a pixel $q$ for the surface point ${\bf M}$. 
The angle between the incident ray and the reflected ray (visual ray) will restrict the surface point that can be reconstructed by our method. Given a reference plane ${\bf XY}$ as in Fig.~\ref{fig:angle_analysis}, the minimum angle between the incident ray and the reflected ray is $\angle {\bf CMX}$ leading to a surface normal $\overrightharpoon{\bf MU}$, while the maximum angle is $\angle {\bf CMY}$ leading to a surface normal $\overrightharpoon{\bf MV}$. The surface point ${\bf M}$ can be reconstructed by our method when its normal lies in the range between $\overrightharpoon{\bf MU}$ and $\overrightharpoon{\bf MV}$. This range can be represented as the angle $\Delta = \angle {\bf UMV}$, which is related to the distance between the surface and the reference plane, the size of the reference plane, and the relative pose between the camera and the reference plane.

Let $\angle {\bf CMO} = \theta$, $\|\overrightharpoon{\bf MO}\| = h$, $\|\overrightharpoon{\bf XO}\| = w_1$, and $\|\overrightharpoon{\bf YO}\| = w_2$. According to the law of reflection, we have
\begin{gather}
\angle {\bf CMU}= \frac{1}{2}(\theta - tan^{-1} \frac{w_1}{h}), \label{eq:angle_u} \\
\angle {\bf CMV}= \frac{1}{2}(\theta + tan^{-1} \frac{w_2}{h}). \label{eq:angle_v} 
\end{gather}
It follows that 
\begin{equation}
\begin{aligned}
\Delta &= \angle {\bf CMV} - \angle {\bf CMU}\\
       &= tan^{-1} \frac{w_1}{h} + tan^{-1} \frac{w_2}{h}. \label{eq:angle_delta}
\end{aligned}
\end{equation}
From (\ref{eq:angle_delta}), we can see that a larger size of the reference plane and/or a closer distance between the surface and the reference plane will result in a larger surface region that can be reconstructed by our method. Note that however according to (\ref{eq:angle_u}), it is not necessary for $w_1$ to be infinite large as we have to ensure $\angle {\bf CMU} \geq 0$ (i.e., $w_1 \leq h \times tan \theta$ ) otherwise the camera will not perceive the reflected ray $\overrightharpoon{\bf CM}$.

Thus, we can conclude that if the surface normals are distributed within the range defined by $\Delta$, the whole surface can be reconstructed by our method. For a surface with a broader normal distribution, we can recover it by increasing the $\Delta$ range via using a large enough reference plane and placing the reference plane at a location not too far away from the surface.
In addition, placing the reference plane at multiple locations facing different (visible) regions of the object can also help enlarge the region that can be reconstructed. Alternatively, a cubic room like the one built in \cite{Balzeretal2014} can be used to enlarge the normal range that can be reconstructed by our method. However, our method requires moving each side of the room to three different locations, which makes it difficult to build such a room.

\begin{figure} [htbp]
   \centering
  \includegraphics[width=0.5\linewidth]{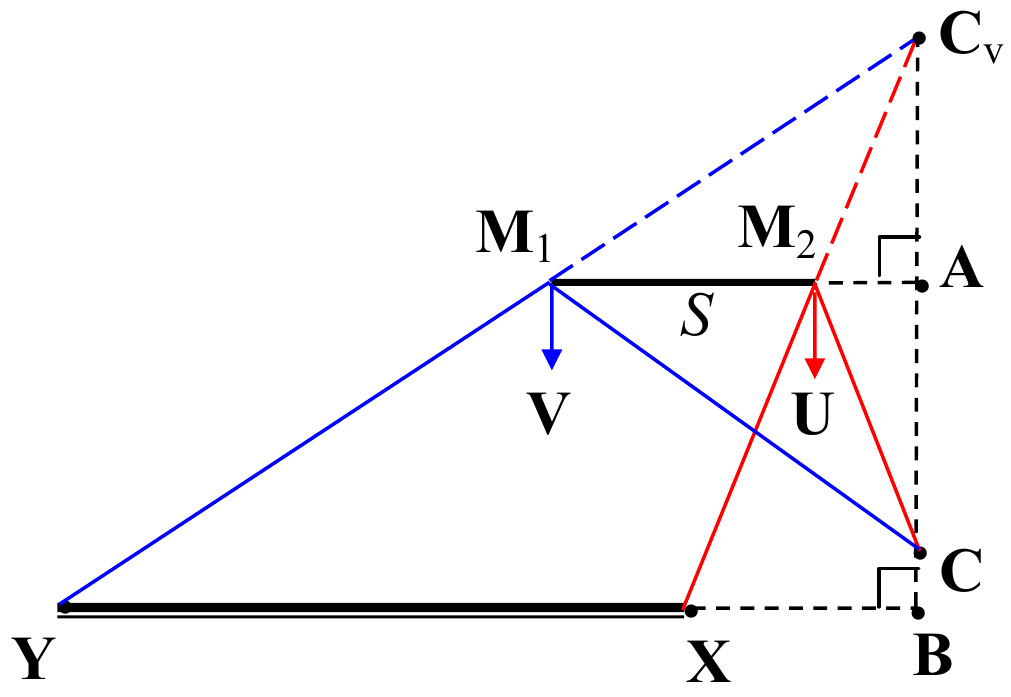}
   \caption{The relationship among the object size, distance from the camera to the object, and distance from the reference plane to the object.}
   \label{fig:mirror_dist}
\end{figure}

In addition, we discuss the relationship among the object size, distance from the camera to the object, and distance from the reference plane to the object as shown in~\fref{fig:mirror_dist}. A camera centered at $\bf C$ observes the mirror surface $S$. 
$\overrightharpoon{{\bf M}_1 {\bf V}}$ and $\overrightharpoon{{\bf M}_2 {\bf U}}$ are the normals at surface points ${\bf M}_1$ and ${\bf M}_2$.
For simplicity, let $\overrightharpoon{{\bf M}_1 {\bf M}_2} \parallel \overrightharpoon{{\bf X} {\bf Y}}$.
We then construct points $\bf A$ and $\bf B$ with $\overrightharpoon{\bf C \bf A} \perp \overrightharpoon{{\bf M}_1 {\bf M}_2}$ and $\overrightharpoon{\bf A \bf B} \perp \overrightharpoon{\bf X \bf Y}$.
A virtual camera center at ${\bf C}_v$ can be formed by extending $\overrightharpoon{{\bf Y}{\bf M}_1}$ and $\overrightharpoon{{\bf X} {\bf M}_2}$. Let $\|\overrightharpoon{{\bf C}{\bf A}}\|=h_1$, $\|\overrightharpoon{{\bf A}{\bf B}}\|=h_2$, $\|\overrightharpoon{{\bf M}_1{\bf M}_2}\|=s_1$, and $\|\overrightharpoon{{\bf X}{\bf Y}}\|=s_2$, we have $\frac{\|\overrightharpoon{{\bf C}_v \bf A}\|}{\|\overrightharpoon{{\bf C}_v \bf A}\| + \| \overrightharpoon{\bf A \bf B} \|} = \frac{\|\overrightharpoon{\bf C \bf A}\|}{\|\overrightharpoon{\bf C \bf A}\| + \| \overrightharpoon{\bf A \bf B} \|} = \frac{h_1}{h_1+h_2} = \frac{\|\overrightharpoon{{\bf M}_1{\bf M}_2}\|}{\|\overrightharpoon{\bf X \bf Y}\|} = \frac{s_1}{s_2}$. The maximum size of the surface that can be reconstructed by our method is then approximated by $s_1 = \frac{h_1s_2}{h_1+h_2}$. Note that we use a planar surface $S$ here only for discussion, while planar surfaces correspond to the degenerate case of our method as will be discussed next. As long as there are some normal variations between ${\bf M}_1$ and ${\bf M}_2$ on $S$, the above discussion is still a valid approximation.

\begin{figure}[htb]
\centering 
\includegraphics[width=0.6\linewidth]{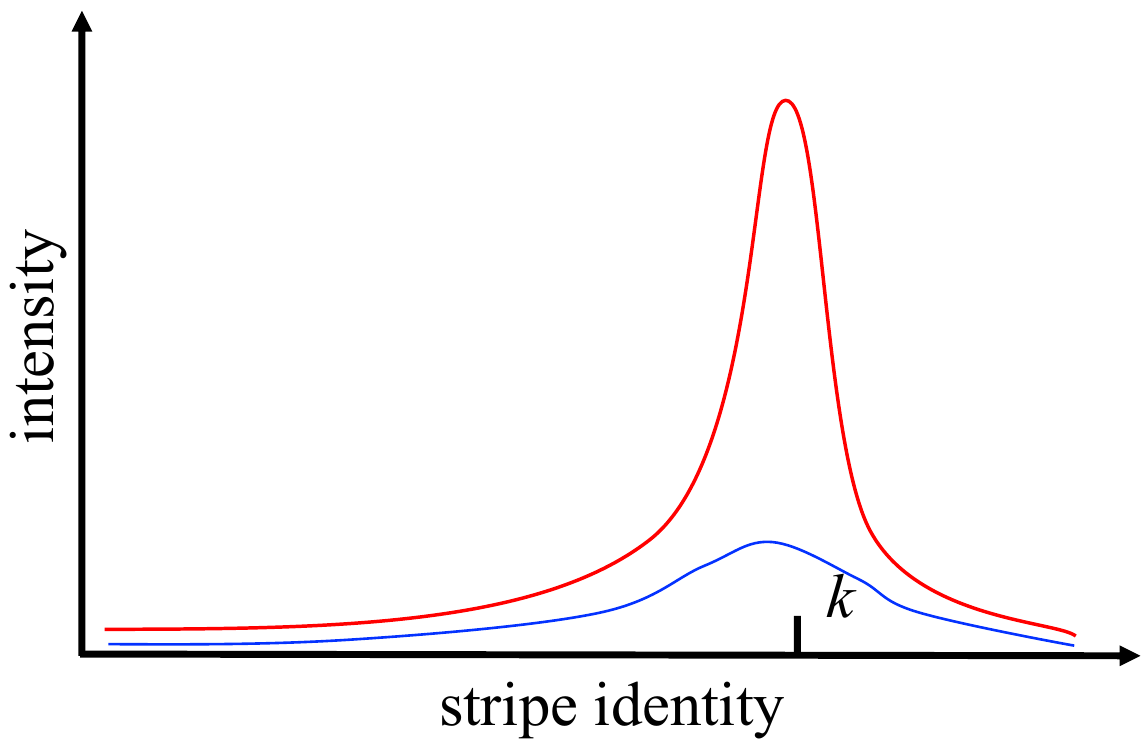}
\caption{An example of intensity profile. The red curve denotes the case that all energy of the incident ray is reflected. Thus the peak (i.e., $k$-th stripe) is sharper and easier to identify (small uncertainty). While the blue cure denotes the case that only part of the energy is reflected, resulting in a relatively flat profile, whose peak is more difficult to determine (large uncertainty).}
\label{fig:stripe}
\end{figure}

Lastly, it is worth noting that we assume the surface is perfectly reflective and our formulation depends on the law of reflection. Therefore, our method may not work for other mirror-like surfaces that do not follow the law of reflection. 
Meanwhile, for some mirror objects, the energy of the incident rays are not fully reflected due to the surface material. For such cases, the brightness of the sweeping stripes for correspondence estimation will be low, bringing more difficulties in identifying the peak from the intensity profile (see~\fref{fig:stripe}), thus reducing the accuracy of camera projection matrix estimation and shape reconstruction.

\subsection{Degeneracy}
In order to obtain 3D line correspondences, we proposed the method in Section~\ref{sec:planepose} to estimate relative poses of the reference plane. However, there do exist degenerate cases. 
In particular, the relative poses of the reference plane cannot be uniquely determined in the following cases:
(1)  The specular object is of a planar, elliptical, parabolic or hyperbolic mirror;  and
(2)  the arrangement of such mirror surfaces and the pinhole camera forms a central catadioptric system; namely the pinhole camera and the mirror surfaces form a single effective viewpoint. If the unknown mirror surface and the camera form a central catadioptric system with a single effective viewpoint, the relative pose estimation between reference planes (e.g., $P_0$, $P_1$, $P_2$) is equivalent to the camera calibration process described in~\cite{Zhang1998techreport}. As demonstrated in~\cite{Zhang1998techreport}, the poses cannot be determined if the reference plane undergoes pure translation without knowing the translation direction. If the reference plane is placed at three general positions, the poses are estimated up to one scale ambiguity. As analysed in~\cite{Baker1999ijcv}, a pinhole camera viewing a sphere does not form a central catadioptric system in general if the distance between the pinhole and effective viewpoint is larger than the radius.
If the camera is located at the centre of the sphere, it indeed forms a central catadioptric system, which however differs from our experimental setup.

In addition, when $\{{\bf m}, {\bf x}_0; {\bf x}_1, {\bf x}_2\}$ in Fig.~\ref{fig:cross_ratio} are close to each other, while $\{{\bf M}, {\bf X}_0; {\bf X}_1, {\bf X}_2\}$ are further away to each other, our cross-ratio based formulation will be more sensitive to noise and a small error in the pixel domain will result in a large error in 3D points estimation. Actually, this is on par with small baseline triangulation as described in \cite{hartleyMVG}, which is generally more sensitive to noise.

\section{Evaluation}
\label{evaluation}
To demonstrate the effectiveness of our method, we evaluate it using both synthetic and real data. 
We first evaluate the effectiveness of our proposed approach for estimating the relative poses of the reference plane (Section \ref{sec:planepose}). We then show that our approach can faithfully reconstruct the mirror surfaces. 
The source code of our method can be found at~\url{https://github.com/k-han/mirror}.

\subsection{Relative Pose Estimation of the Reference Plane}
\begin{figure}[htbp]
\begin{center}
\tabcolsep=0.03cm
   \renewcommand{\arraystretch}{0.01}
  \begin{tabular}{
    >{\centering\arraybackslash} m{0.16\textwidth}
    >{\centering\arraybackslash} m{0.16\textwidth}
    >{\centering\arraybackslash} m{0.16\textwidth}}
  \includegraphics[width=\linewidth]{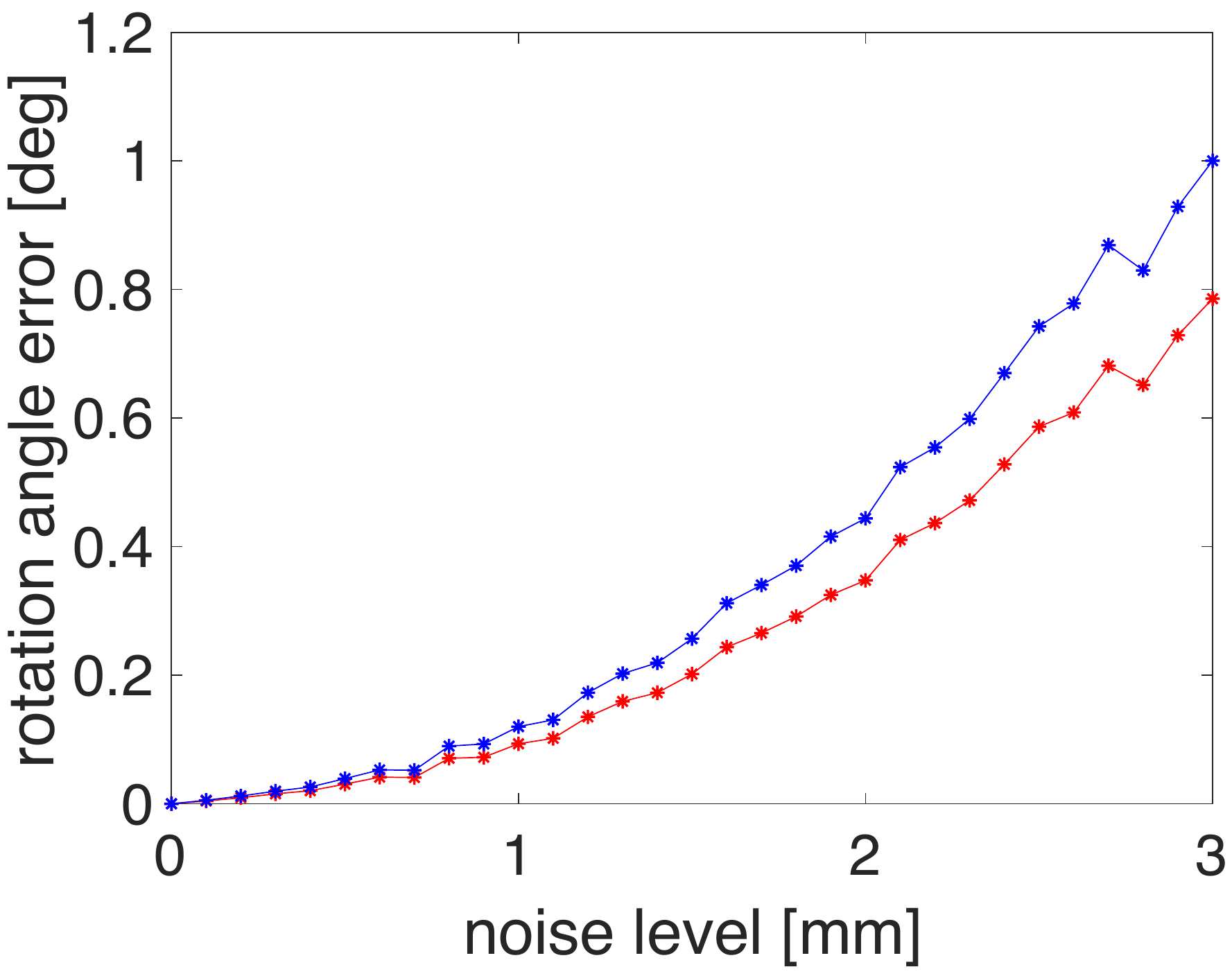} &
  \includegraphics[width=\linewidth]{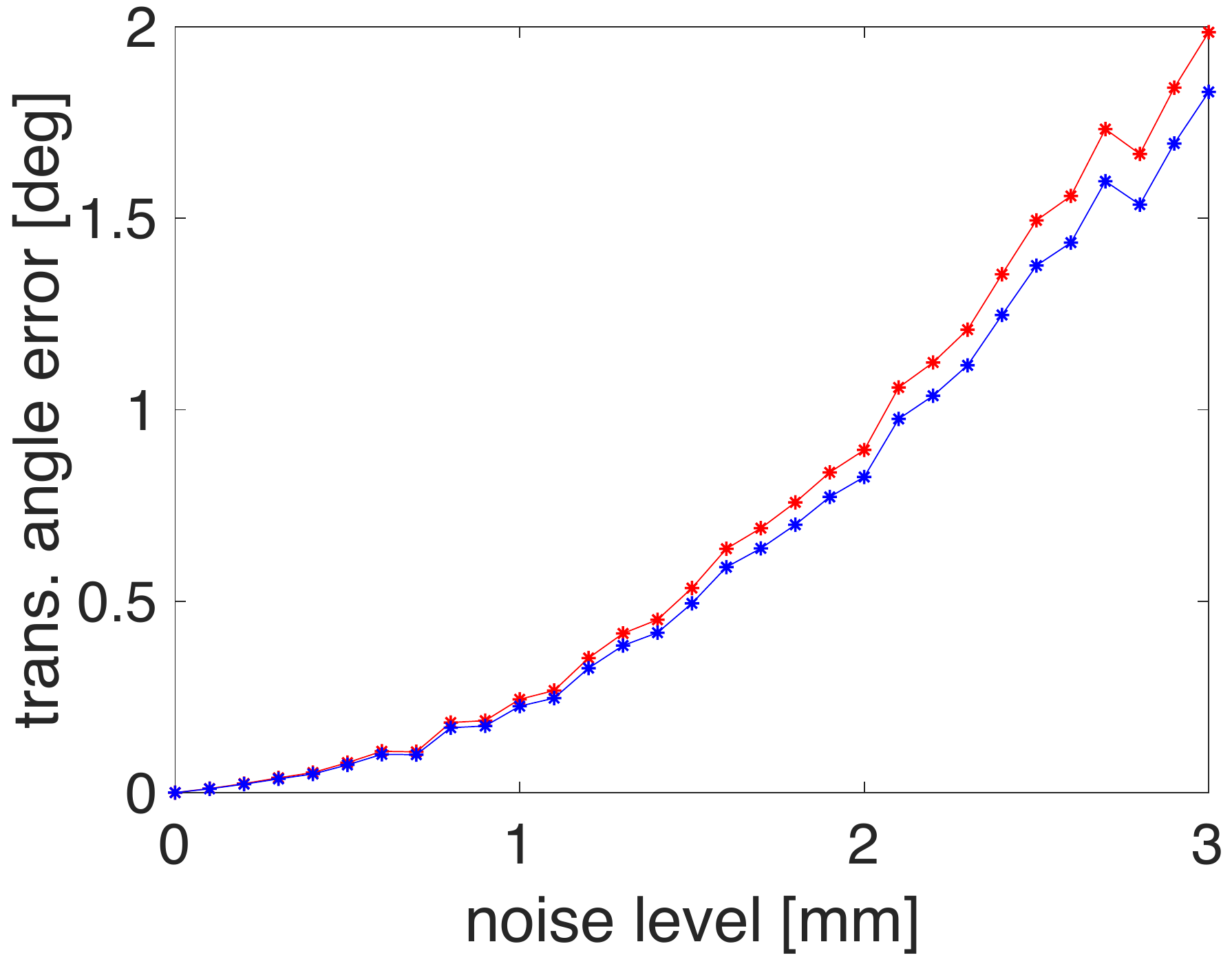} &
  \includegraphics[width=\linewidth]{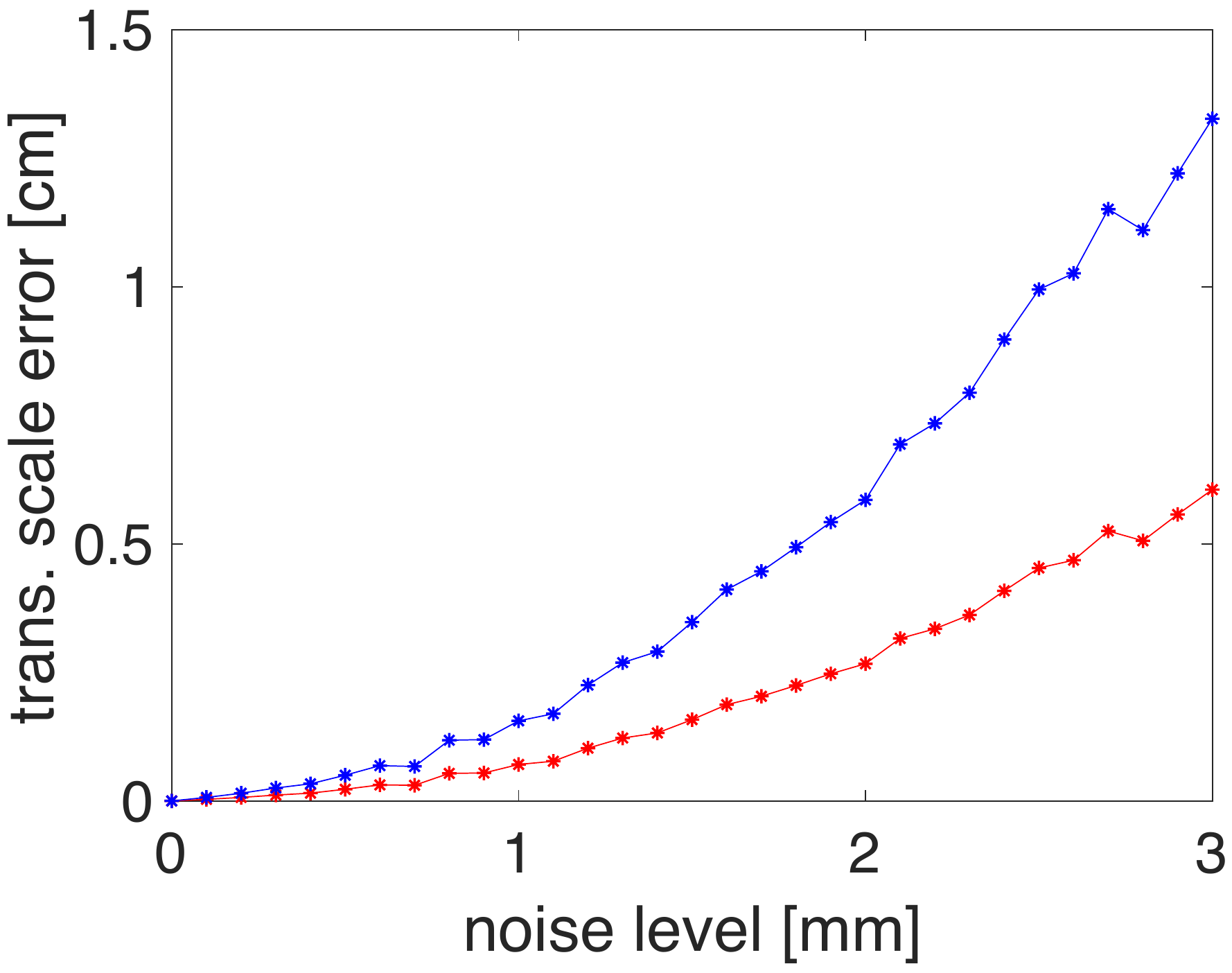}\\
  (a) & (b) & (c)
\end{tabular}
\end{center}
\caption{Noise sensitivity analysis for relative pose estimation of the reference plane. \textmd{The~\emph{red curves} and~\emph{blue curves} are the pose estimation errors for $P_1$ and $P_2$, respectively, under different noise levels. (a) Relative rotation error. (b) Translation angular error. (c) Translation scale error. }}
\label{fig:poseErr}
\end{figure}

To evaluate the performance of our reference plane pose estimation presented in Section~\ref{sec:planepose}, we generated synthetic data using two spheres. The spheres have a radius of 300 $mm$. The reference plane used has a dimension of 2000$\times$2000 $mm^2$ and was placed at three different poses denoted by $P_0$, $P_1$, and $P_2$ respectively. The reflection correspondences were obtained via ray tracing. To evaluate the robustness of our method, we added Gaussian noise to the reflection correspondences on the reference plane with standard deviations ranging from 0 to 3.0 $mm$. The errors were reported as the average value over 50 trials of the experiments for each noise level (Gaussian noise standard deviation). Specifically, we reported the errors in the relative rotation matrix ${\bf R}$ in terms of the angle of the rotation induced by ${\bf R}_{gt}{\bf R}^{\rm T}$, where ${\bf R}_{gt}$ denotes the ground truth rotation matrix. We reported the errors in the translation vector ${\bf T}$ in terms of the angle (${\bf T}_{deg}$) between ${\bf T}$ and ${\bf T}_{gt}$, where ${\bf T}_{gt}$ denotes the ground truth translation vector and ${\bf T}_{scale} = \|{\bf T}_{gt} - {\bf T}\|$. In Fig.~\ref{fig:poseErr}, we show the estimation errors for (${\bf R}^1$, ${\bf T}^1$) and (${\bf R}^2$, ${\bf T}^2$). It can be seen that the errors increase with the noise level, while the magnitude of the errors are quite small, demonstrating that our relative pose estimation method is robust to the noise.

\subsection{Recovery of Projection Matrix and Mirror Surface}
\subsubsection{Synthetic Data Experiments}

\begin{figure}[htbp]
\begin{center}
\tabcolsep=0.03cm
   \renewcommand{\arraystretch}{0.01}
   \begin{tabular}{
         >{\centering\arraybackslash} m{0.1\textwidth}
         >{\centering\arraybackslash} m{0.18\textwidth}
         >{\centering\arraybackslash} m{0.18\textwidth}}
      \includegraphics[width=\linewidth]{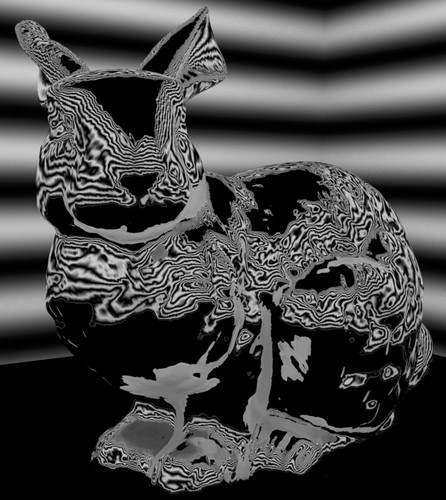} &
      \includegraphics[width=\linewidth]{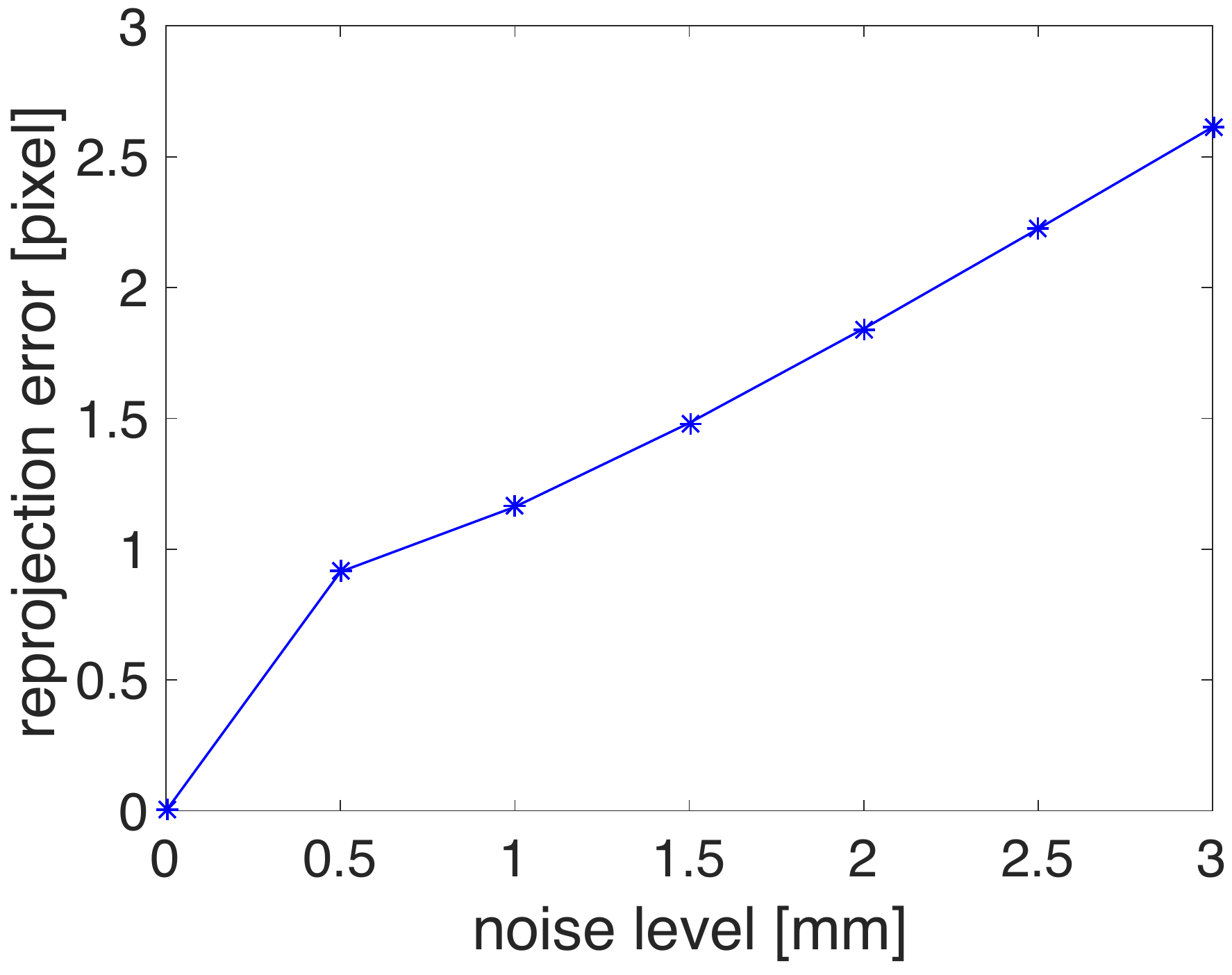} &
      \includegraphics[width=\linewidth]{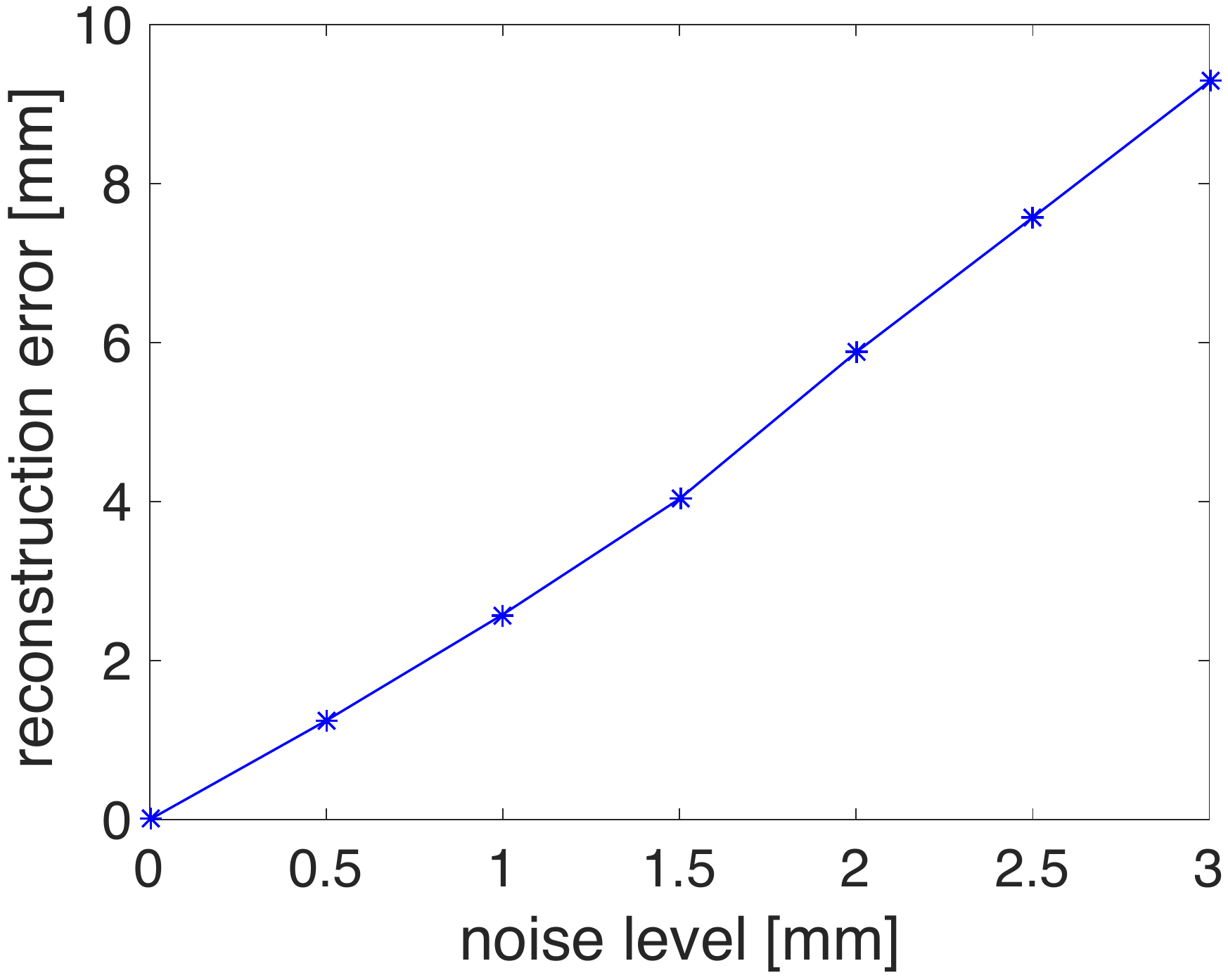} \\
      \includegraphics[width=\linewidth]{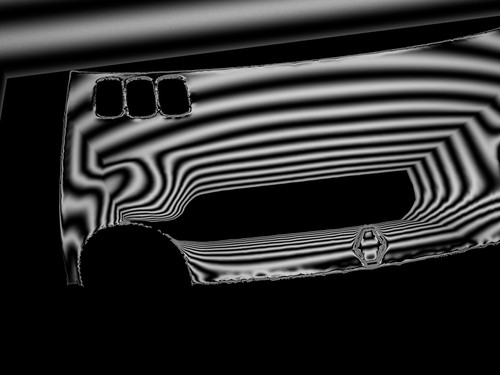} &
      \includegraphics[width=\linewidth]{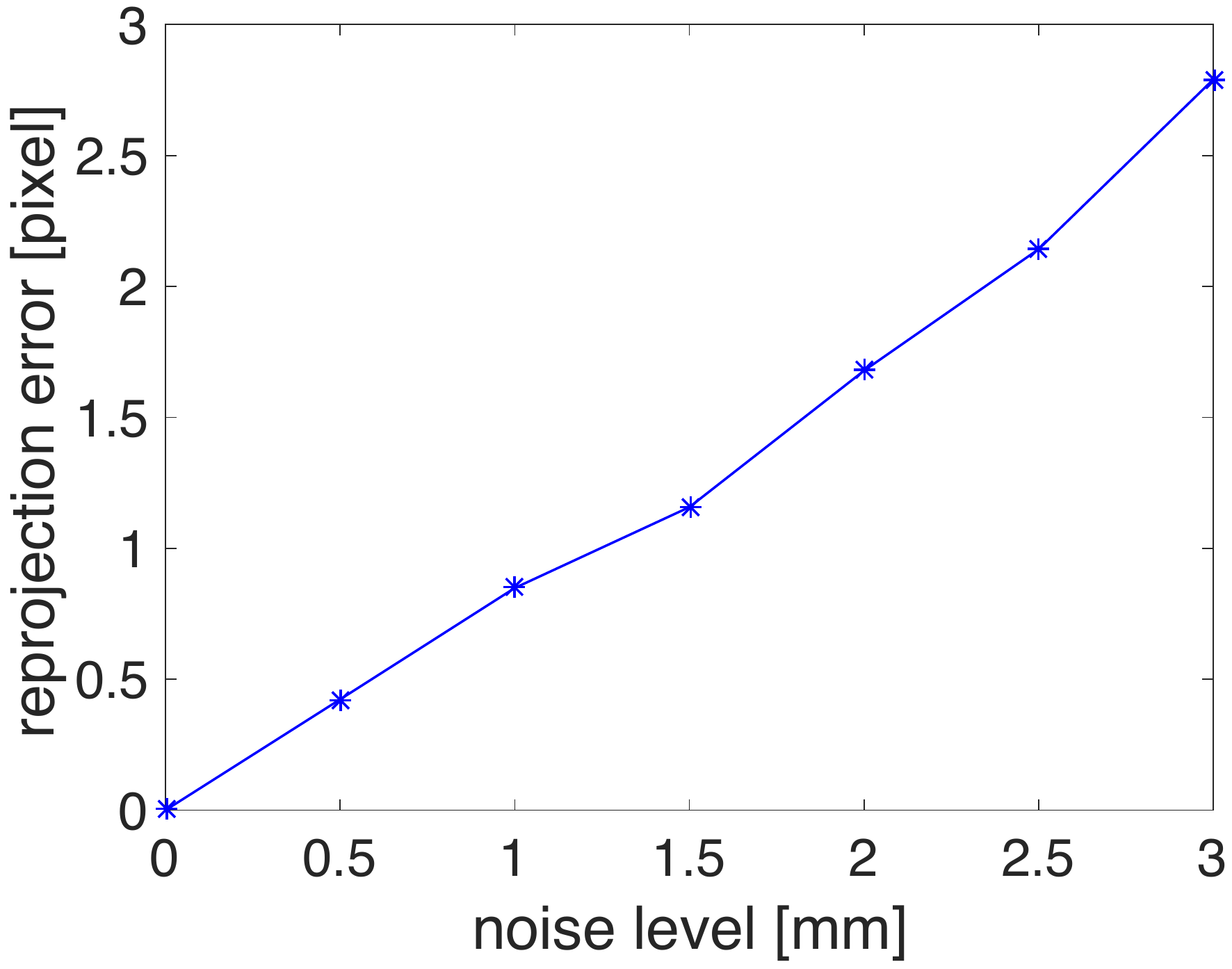} &
      \includegraphics[width=\linewidth]{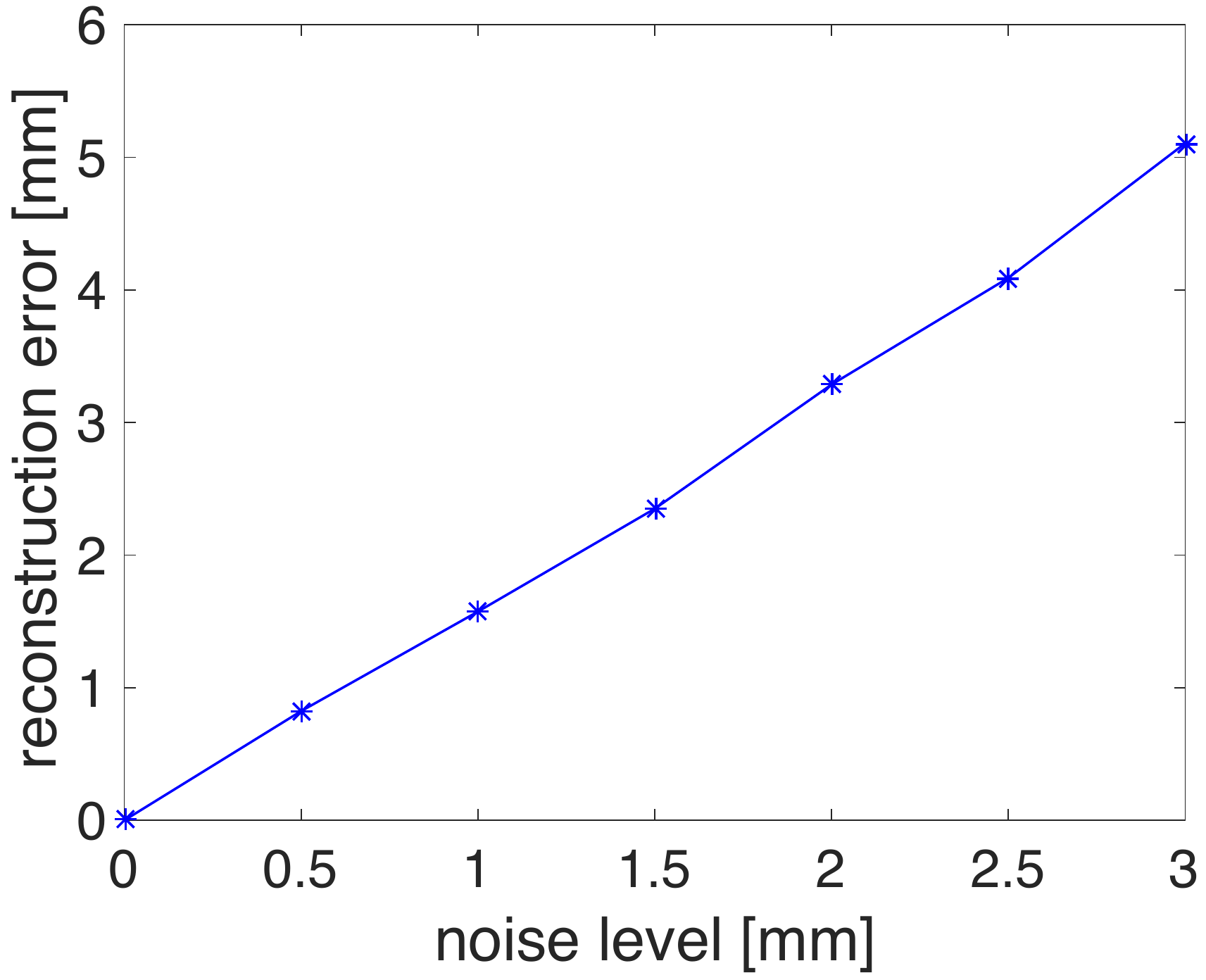} \\
      (a) &
      (b) &
      (c)\\
   \end{tabular}
\end{center}
   \caption{RMS errors for the mirror~\emph{Stanford bunny} (first row) and \emph{engine hood} (second row). \textmd{(a) An image of the mirror~\emph{Stanford bunny} and an image of the mirror~\emph{engine hood}. (b) RMS reprojection errors (computed against ground truth image points). (c) RMS reconstruction errors (computed against ground truth 3D surface points).}}
   \label{fig:bunny_rms}
\end{figure}

We employed a reflective~\emph{Stanford bunny} and a reflective~\emph{engine hood} created by \cite{Balzeretal2014} to generate our synthetic data. The bunny has a dimension of $880\times680\times870$ $mm^3$ and $208,573$ surface points, while the hood has a dimension of $2120\times1180\times270$ $mm^3$ and $38,546$ surface points. The images have a resolution of $960\times 1280$ pixels. \Fref{fig:bunny_rms}(a) shows the reflective appearances of the bunny and the hood. In their original data, the bunny (hood) was placed in a cubic room, with each side of the room working as a reference plane. The reference plane has a dimension of $3048\times3048$ $mm^2$. The center of the room was defined as the world origin. A camera was placed in the room viewing the bunny (hood). Since our method requires reflection correspondences under three distinct poses of the reference plane, we introduced two additional planes for each side of the room and obtained the reflection correspondences through ray tracing.

\begin{table}[htbp]
\centering
   \tabcolsep=3.5pt
   \renewcommand{\arraystretch}{1.0}
   \caption{Estimation error under noise lv $\sigma = 2.0$ $[mm]$ on $bunny$.}
   \label{tab:linear_p}
   \begin{threeparttable}
   \begin{tabular}{l | c c c c c c c}
   \hline
      &  $f_u\;[\%]$ & $f_v\;[\%]$ & $u_0\;[\%]$ & $v_0\;[\%]$ & $\bf R\;[^{\circ}]$ & ${\bf T}_{deg}\;[^{\circ}]$ & ${\bf T}_{scale}\;[\%]$ \\
      \hline
      $L$ &  $1.39$ & $1.78$ & $1.76$ & $2.39$ & $0.61$ & $0.55$ & $0.98$ \\  
      $EL$ &  $0.94$ & $0.94$ & $0.07$ & $0.10$ & $0.12$ & $0.11$ & $0.28$ \\  
      $CR$ &  $0.11$ & $0.11$ & $0.18$ & $0.25$ & $0.08$ & $0.07$ & $0.13$ \\ 
   \hline
   \end{tabular}
    \begin{tablenotes}
      \item  \textmd{$L$: linear solution in Section~\ref{initialization}; $EL$: constrained linear solution with strategy in Section~\ref{initialization_E}; $CR$: estimation using cross-ratio formulation initialized with $EL$.}
    \end{tablenotes}
   \end{threeparttable}
\end{table}

\begin{figure*}[htbp]
  \centering
   \tabcolsep=0.02cm
   \begin{tabular}{
         >{\centering\arraybackslash} m{0.195\textwidth}
         >{\centering\arraybackslash} m{0.195\textwidth}
         >{\centering\arraybackslash} m{0.195\textwidth}
         >{\centering\arraybackslash} m{0.195\textwidth}
         >{\centering\arraybackslash} m{0.195\textwidth}}
      \includegraphics[width=1\linewidth]{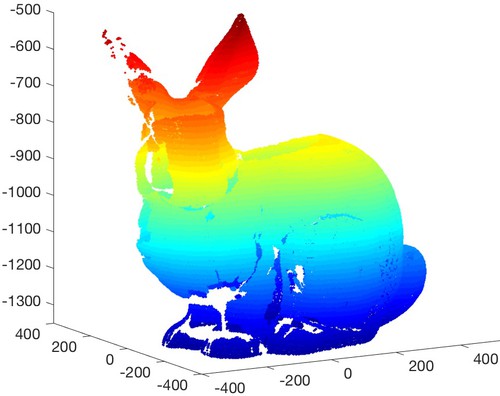} &
      \includegraphics[width=1\linewidth]{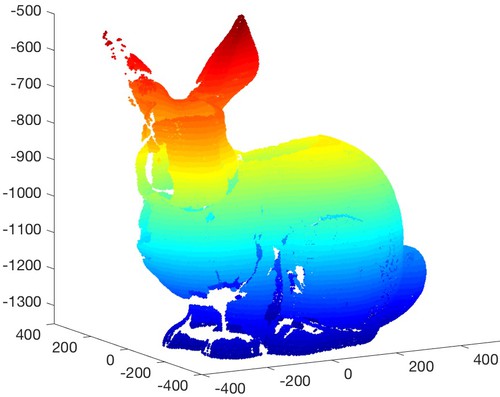} &
      \includegraphics[width=1\linewidth]{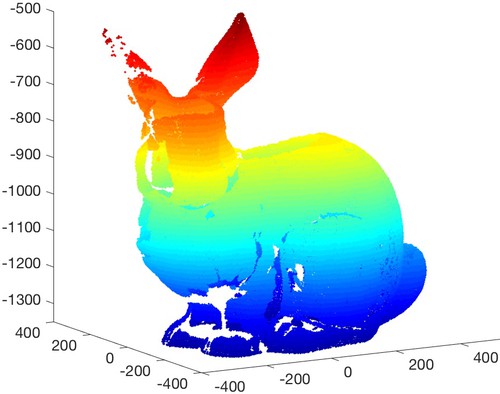} &
      \includegraphics[width=1\linewidth]{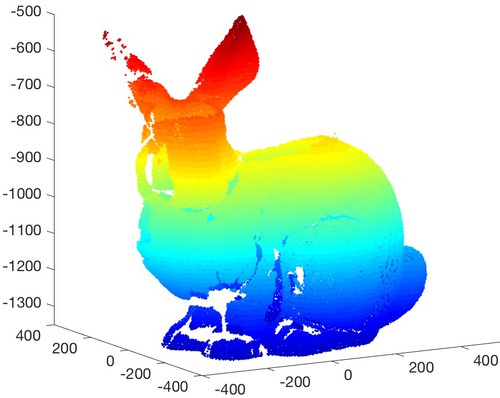} &
      \includegraphics[width=1\linewidth]{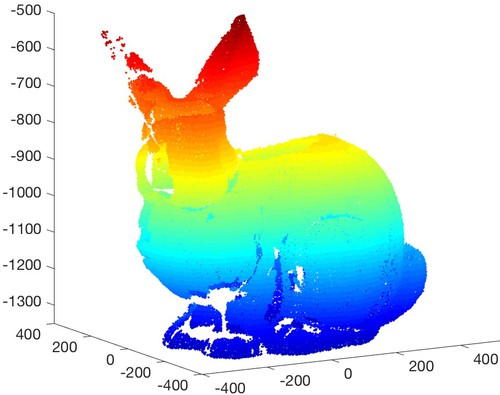} \\
      \includegraphics[width=0.8\linewidth]{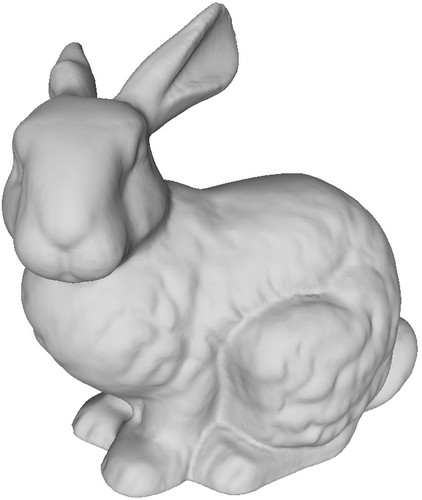} &
      \includegraphics[width=0.8\linewidth]{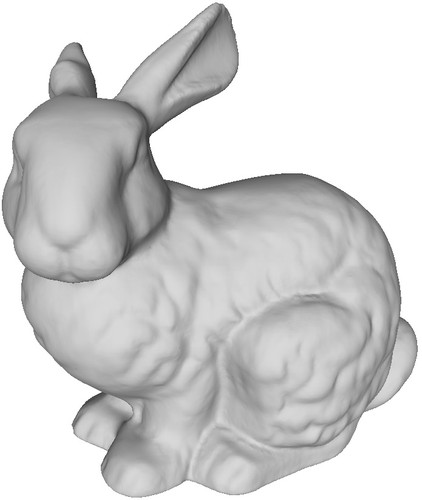} &
      \includegraphics[width=0.8\linewidth]{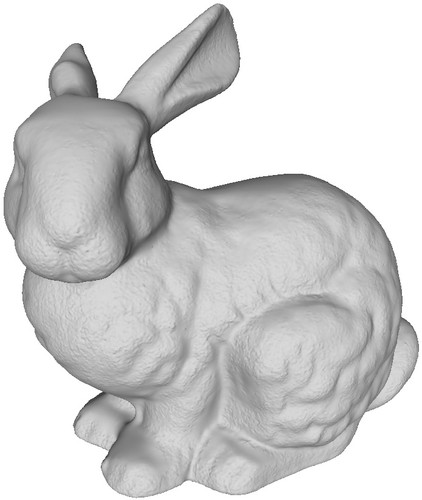} &
      \includegraphics[width=0.8\linewidth]{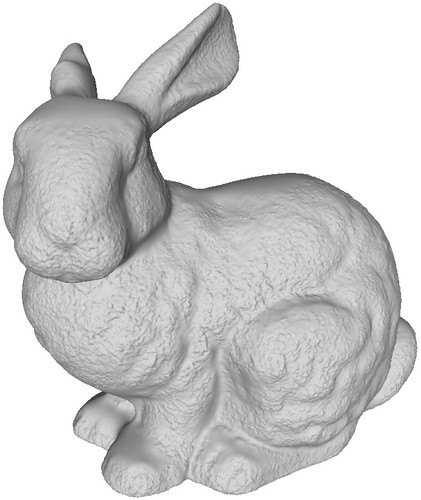} &
      \includegraphics[width=0.8\linewidth]{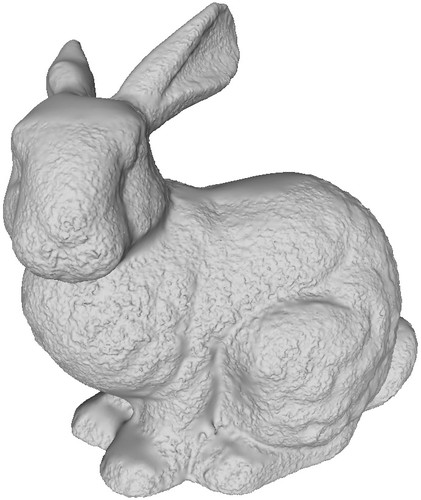} \\
      &
      \includegraphics[width=1\linewidth]{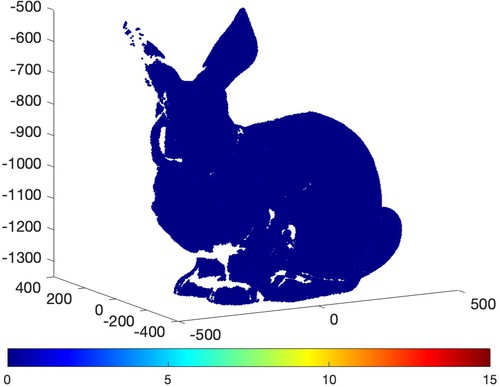} &
      \includegraphics[width=1\linewidth]{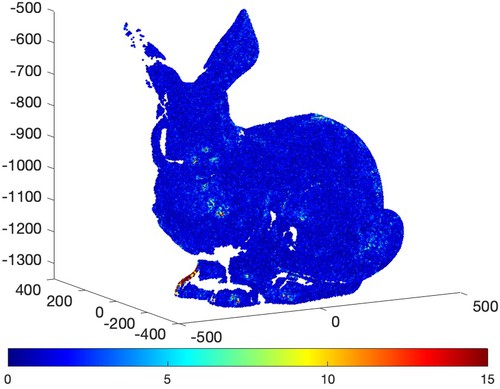} &
      \includegraphics[width=1\linewidth]{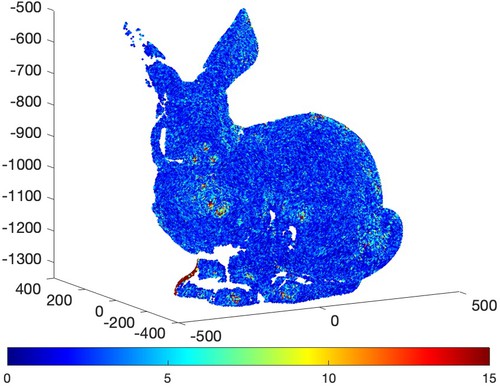} &
      \includegraphics[width=1\linewidth]{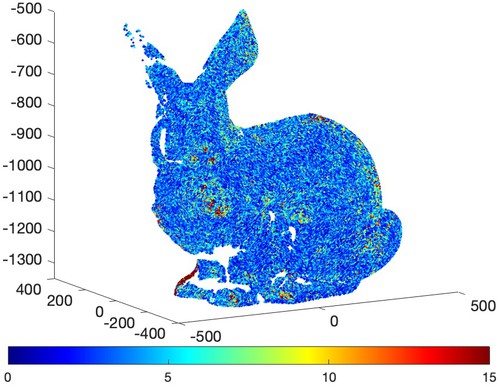} \\
      \includegraphics[width=1\linewidth]{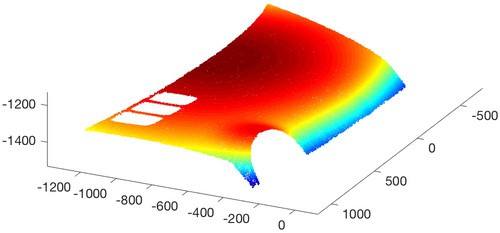} &
      \includegraphics[width=1\linewidth]{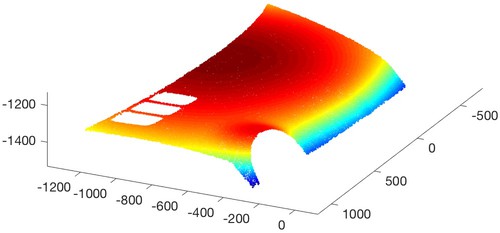} &
      \includegraphics[width=1\linewidth]{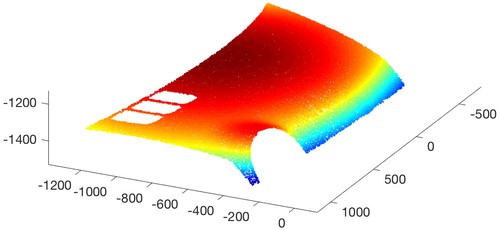} &
      \includegraphics[width=1\linewidth]{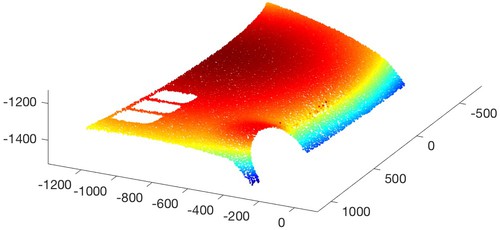} &
      \includegraphics[width=1\linewidth]{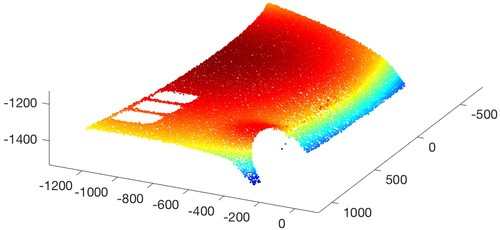} \\
      \includegraphics[width=0.8\linewidth]{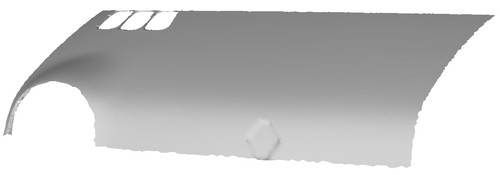} &
      \includegraphics[width=0.8\linewidth]{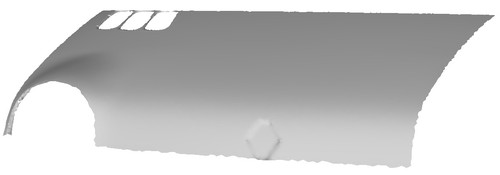} &
      \includegraphics[width=0.8\linewidth]{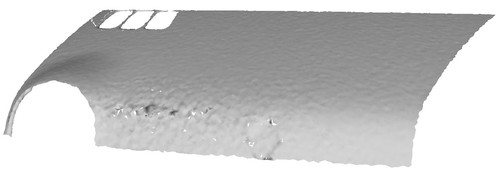} &
      \includegraphics[width=0.8\linewidth]{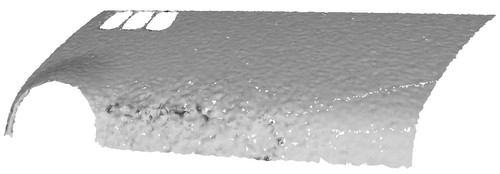} &
      \includegraphics[width=0.8\linewidth]{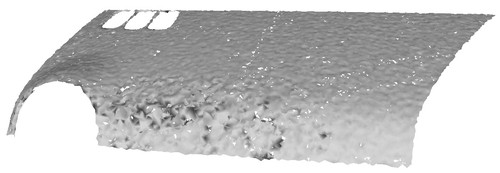} \\
      &
      \includegraphics[width=0.95\linewidth]{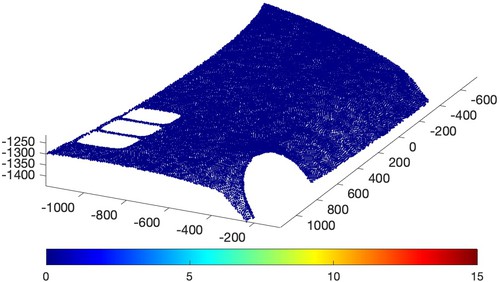} &
      \includegraphics[width=0.95\linewidth]{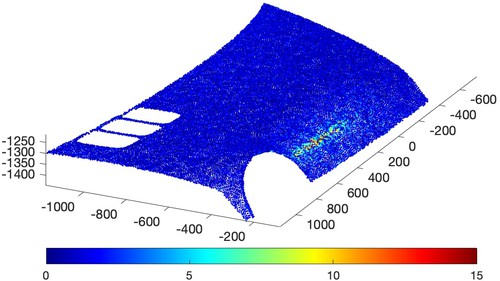} &
      \includegraphics[width=0.95\linewidth]{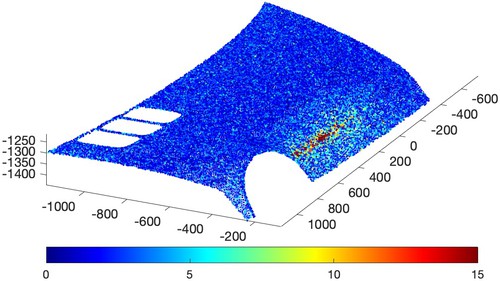} &
      \includegraphics[width=0.95\linewidth]{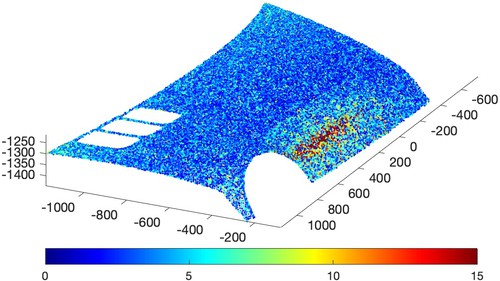} \\
      ground truth &
      no noise &
      noise lv: $\sigma = 1.0$ &
      noise lv: $\sigma = 2.0$ &
      noise lv: $\sigma = 3.0$\\
   \end{tabular}
   \caption{Reconstruction results of \emph{bunny} and \emph{hood} shapes. \textmd{First and fourth rows: reconstructed point clouds under different noise levels of the bunny and the hood, respectively. Coordinates are w.r.t world and colors are rendered w.r.t $z$ coordinates. Note that the missing regions of the bunny are due to the lack of correspondences in the original data set. Second and fifth rows: surfaces of the bunny and the hood generated using screened Poisson surface reconstruction method in \cite{kazhdan2013tog}. Third and last rows: visualization of surface point reconstruction errors.}}
   \label{fig:bunny_recons}
\end{figure*}

\begin{table*}[htbp]
\centering
   \tabcolsep=7pt
   \renewcommand{\arraystretch}{1.0}
   \caption{Camera intrinsic and extrinsic estimation errors under different noise levels $\sigma$ for \emph{Stanford bunny} dataset and \emph{engine hood} dataset. }
   \label{tab:synexp}
   \begin{threeparttable}
   \begin{tabular}{l | l | c c c c c c c}
   \hline
      &$noise\;lv$&  $f_u\;[pixel]$ & $f_v\;[pixel]$ & $u_0\;[pixel]$ & $v_0\;[pixel]$ & $\bf R\;[^{\circ}]$ & ${\bf T}_{deg}\;[^{\circ}]$ & ${\bf T}_{scale}\;[mm]$ \\
      \hline
      \multirow{6}{*}{\rotatebox[origin=c]{90}{bunny}}& $\sigma = 0.5$ &  $0.22(0.02\%)$ & $0.22(0.02\%)$ & $0.41(0.06\%)$ & $0.07(0.01\%)$ & $0.02$ & $0.02$ & $0.53(0.03\%)$ \\
      & $\sigma = 1.0$ &  $0.33(0.02\%)$ & $0.33(0.02\%)$ & $0.41(0.06\%)$ & $0.05(0.01\%)$ & $0.02$ & $0.02$ & $0.52(0.03\%)$ \\  
       & $\sigma = 1.5$ &  $0.50(0.04\%)$ & $0.51(0.04\%)$ & $0.56(0.09\%)$ & $0.90(0.20\%)$ & $0.04$ & $0.05$ & $1.74(0.10\%)$ \\  
      & $\sigma = 2.0$&  $1.52(0.11\%)$ & $1.52(0.11\%)$ & $1.15(0.18\%)$ & $1.22(0.25\%)$ & $0.08$ & $0.07$ & $2.42(0.13\%)$ \\
      & $\sigma = 2.5$&  $4.36(0.31\%)$ & $4.36(0.31\%)$ & $2.11(0.32\%)$ & $2.36(0.49\%)$ & $0.37$ & $0.56$ & $7.33(0.40\%)$ \\
      & $\sigma = 3.0$&  $10.15(0.73\%)$ & $10.11(0.73\%)$ & $5.76(0.90\%)$ & $3.08(0.64\%)$ & $0.85$ & $0.73$ & $13.29(0.73\%)$ \\
      \hline
      \multirow{6}{*}{\rotatebox[origin=c]{90}{hood}}& $\sigma = 0.5$ &  $0.59 (0.04\%)$ & $0.59(0.04\%)$ & $0.18(0.03\%)$ & $0.71(0.15\%)$ & $0.15$ & $0.29$ & $1.07(0.10\%)$ \\
      & $\sigma = 1.0$ &  $1.43 (0.10\%)$ & $1.43(0.10\%)$ & $1.14(0.18\%)$ & $0.72(0.15\%)$ & $0.20$ & $0.26$ & $2.45(0.23\%)$ \\
       & $\sigma = 1.5$ &  $2.53 (0.18\%)$ & $2.56(0.18\%)$ & $1.10(0.17\%)$ & $1.43(0.30\%)$ & $0.17$ & $0.34$ & $5.01(0.47\%)$ \\
      & $\sigma = 2.0$ &  $4.87 (0.35\%)$ & $4.87(0.18\%)$ & $3.08(0.48\%)$ & $2.66(0.55\%)$ & $0.33$ & $0.35$ & $8.38(0.79\%)$ \\
      & $\sigma = 2.5$ &  $7.32 (0.52\%)$ & $7.32(0.52\%)$ & $4.74(0.74\%)$ & $5.37(1.12\%)$ & $0.53$ & $0.55$ & $14.01(1.32\%)$ \\
      & $\sigma = 3.0$ &  $11.53 (0.82\%)$ & $11.53(0.82\%)$ & $6.55(1.02\%)$ & $7.71(1.61\%)$ & $0.54$ & $0.86$ & $17.57(1.66\%)$ \\           
      \hline
   \end{tabular}
    \begin{tablenotes}
      \item  \textmd{The ground truth for the intrinsic parameters are $f_u = 1400$, $f_v = 1400$, and $(u_0, v_0) = (639.5, 479.5)$. The norm of translation vectors are $1811.2$ and $1057.9$ for $bunny$ and $hood$, respectively.}
    \end{tablenotes}
   \end{threeparttable}
\end{table*}

To evaluate the performance of our method, we added Gaussian noise to the 2D reflection correspondences on the reference plane with standard deviations ranging from $0$ to $3.0$ $mm$. 
The reflection correspondences are the input of our method and essential to the performance. It is difficult to obtain high quality reflection correspondences due to the distortion caused by the reflection of the mirror surface. Thus we added the noise to the reflection correspondences. 
The typical intensity profile for the x- or y-coordinate is a bell-shaped curve (see~\fref{fig:stripe} for an example), which has a similar characteristic as Gaussian distribution, hence we choose Gaussian noise to model the possible noises in the 2D correspondences. 
We initialized the projection matrix using the method described in Section~\ref{closeformsolution}. The optimized projection matrix together with the 3D surface points were obtained by minimizing reprojection errors computed based on our cross-ratio formulation. In Table~\ref{tab:linear_p}, we compared: 1) linear solution by solving (\ref{eq:linearsolution}) in Section~\ref{initialization}, denoted as $L$; 2) linear solution by solving (\ref{eq:solveRT}) after enforcing the constraints in Section~\ref{initialization_E}, denoted as $EL$; and 3) cross-ratio based non-linear solution by solving (\ref{eq:non-linear}) in Section~\ref{robustestimation}, denoted as $CR$. It can be seen that our cross-ratio based formulation can effectively improve the linear solutions.
\Fref{fig:bunny_rms}(b) and (c) depict the root mean square (RMS) reprojection errors and reconstruction errors, respectively, under different noise levels. It can be seen that the reprojection errors and the reconstruction errors increase linearly with the noise level. The magnitude of the reconstruction errors is relatively small compared to the size of the object. \Fref{fig:bunny_recons} shows the reconstructed point clouds and surfaces. Table \ref{tab:synexp} shows a quantitative comparison of our estimated projection matrices w.r.t the ground truth. Among all noise levels, the errors are below $1\%$ for $f_u$, $f_v$, $u_0$, $v_0$ and ${\bf T}_{scale}$, and angular errors are below $1^\circ$ for ${\bf R}$ and ${\bf T}$. 

\begin{figure}[htb]
\centering 
\includegraphics[width=0.5\linewidth]{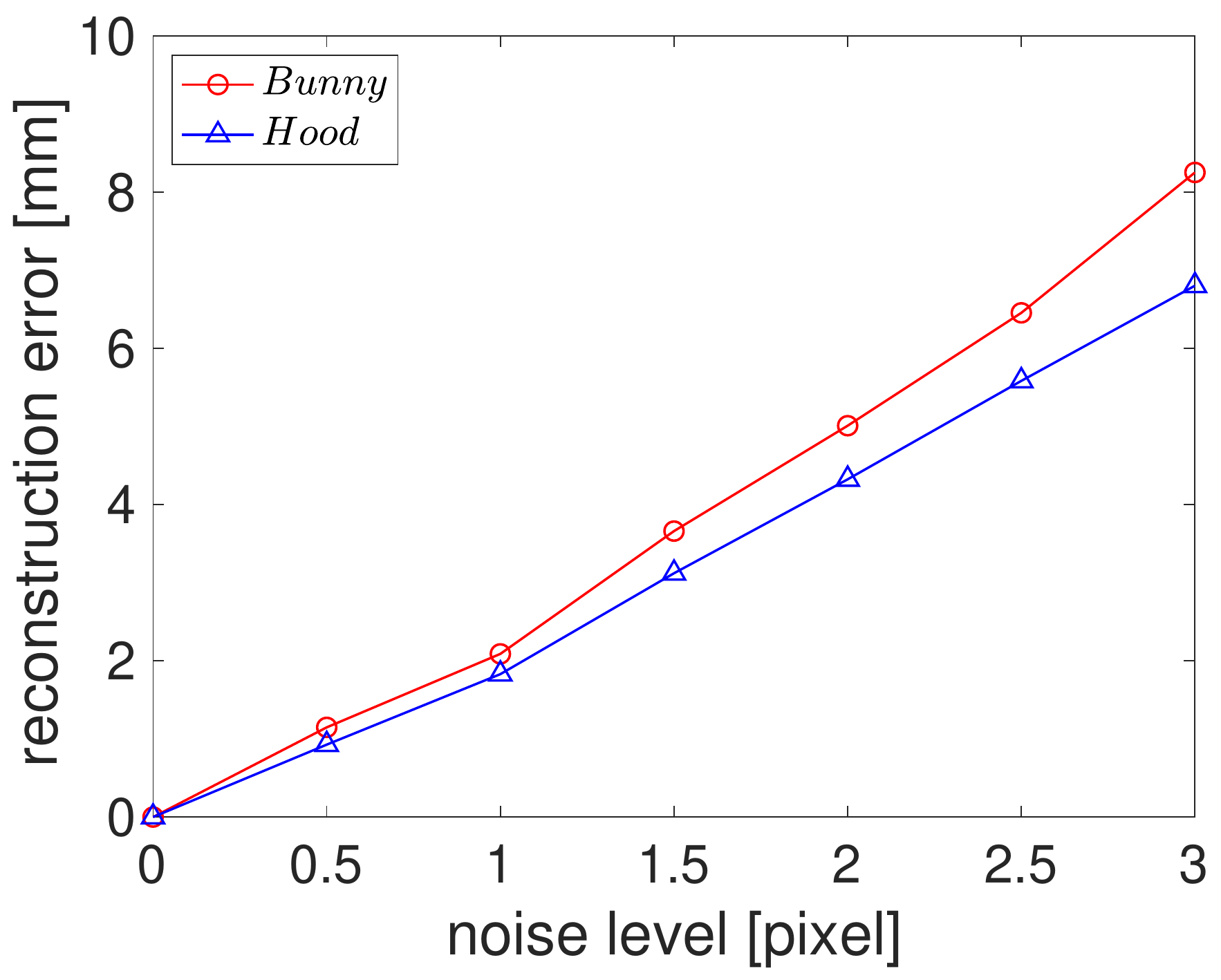}
\caption{Reconstruction results under different quantization noise.}
\label{fig:pix_noise}
\end{figure}

Moreover, as the quantization noise, which obeys the uniform distribution, is also typical in real situations, we further verify the robustness of our approach to the quantization noise by adding the uniform noise in the range of $[-\gamma, \gamma]$ to the 2D pixel locations. The results are reported in~\Tref{tab:quant} and~\fref{fig:pix_noise}. The errors increase linearly with the noise level while the estimation errors are small, revealing that our approach is robust to quantization noise.

\begin{table}[htbp]
   \centering
   \tabcolsep=0.35em
   \renewcommand{\arraystretch}{1.0}
   \caption{Estimation errors on \emph{bunny} and \emph{hood} under quantization noise $\gamma = 2.0$ [$pixel$].}
   \label{tab:quant}
   \begin{threeparttable}
   \begin{tabular}{l | c c c c c c c c}
   \hline
      & $f_u\;[\%]$ & $f_v\;[\%]$ & $u_0\;[\%]$ & $v_0\;[\%]$ & $\bf R\;[^{\circ}]$ & ${\bf T}_{deg}\;[^{\circ}]$ & ${\bf T}_{scale}\;[\%]$ \\
      \hline
      $Bunny$ & $0.12$ & $0.12$ & $0.20$ & $0.72$ & $0.15$ & $0.20$ & $0.36$\\
      $Hood$ & $0.08$ & $0.08$ & $0.11$ & $0.27$ & $0.03$ & $0.07$ & $0.17$\\   
    \hline
   \end{tabular}
   \end{threeparttable}
\end{table}

\begin{figure}[htbp]
  \centering
   \tabcolsep=0.02cm
   \begin{tabular}{
         >{\centering\arraybackslash} m{0.23\textwidth}
         >{\centering\arraybackslash} m{0.25\textwidth}}
      \includegraphics[width=1\linewidth]{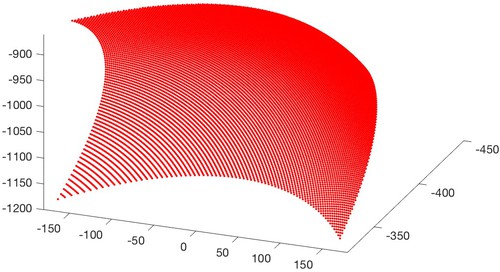} &
      \includegraphics[width=1\linewidth]{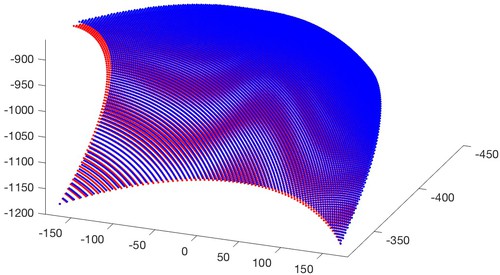}\\
      \includegraphics[width=1\linewidth]{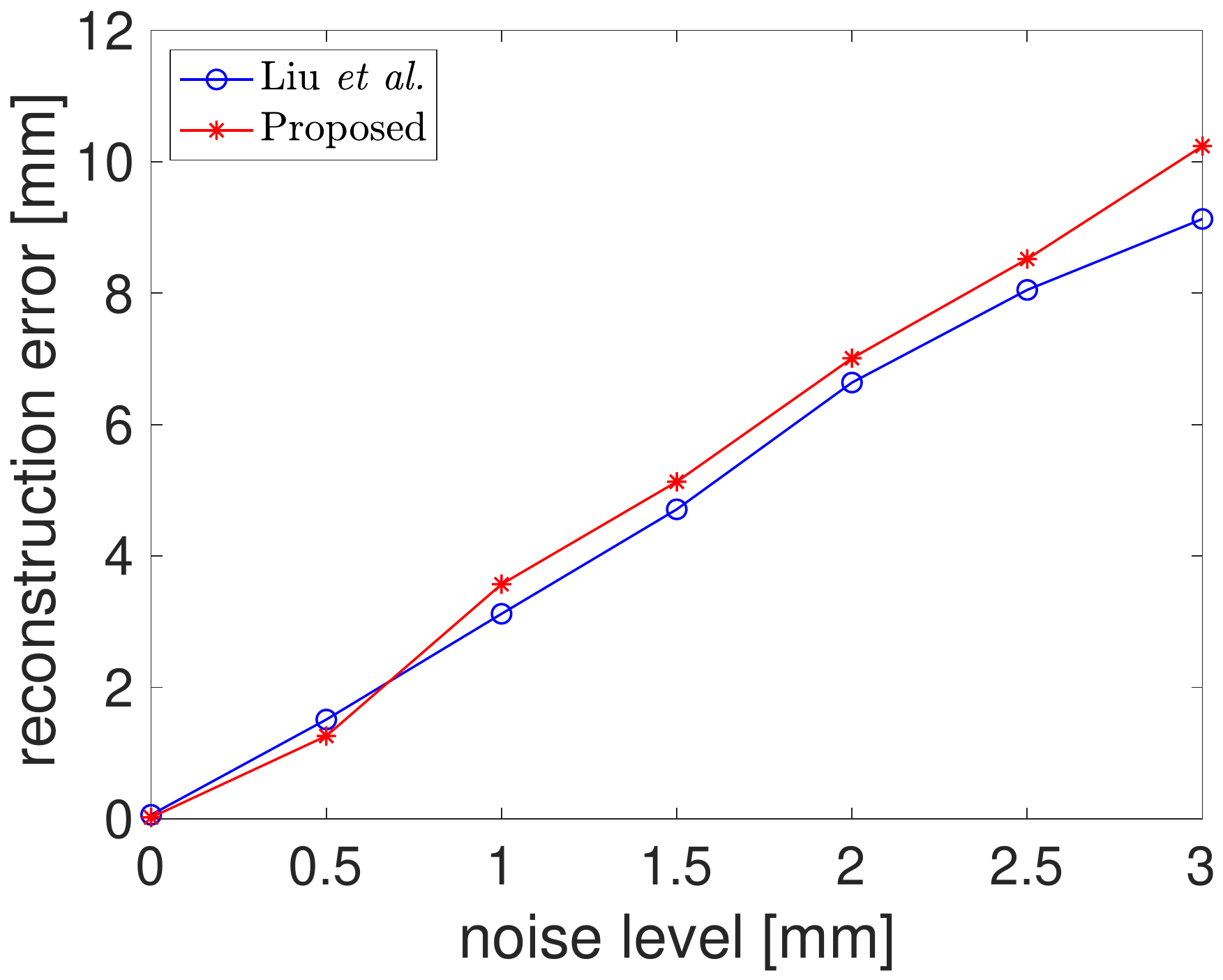}&
      \includegraphics[width=1\linewidth]{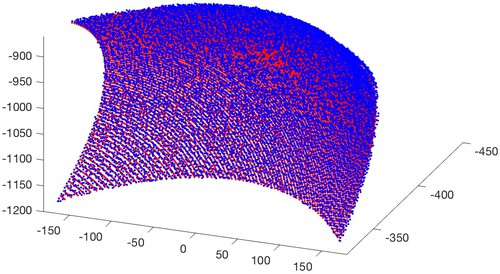}\\
   \end{tabular}
   \caption{Comparison with a fully calibrated method of Liu \emph{et al.} \cite{Liu2015pami}. \textmd{Upper left: ground truth. Lower left: RMS reconstruction errors. Upper right (\cite{Liu2015pami}) \& lower right (ours): reconstruction (blue) against ground truth (red) under $\sigma = 2.0$. Our uncalibrated approach achieves comparable accuracy with that of the fully calibrated method in \cite{Liu2015pami}.}}
   \label{fig:compare}
\end{figure}

Besides, we compared our method with state-of-the-art mirror surface reconstruction method in \cite{Liu2015pami} under smooth surface assumption and calibrated setup. Note that \cite{Liu2015pami} assumes the mirror surface is $C^2$ continuous. In order to make a fair comparison, we performed the experiment on a sphere patch under the same setup with the \emph{Stanford bunny} and \emph{engine hood} datasets. Fig.~\ref{fig:compare} depicts the comparison between fully calibrated (\cite{Liu2015pami}) and uncalibrated (proposed) methods. The overall reconstruction accuracies are similar. While our result is not as smooth as that of \cite{Liu2015pami} due to our point-wise reconstruction. Their result shows a global reconstruction bias due to the B-spline parameterization for the surface (see Fig.~\ref{fig:compare}).

\begin{table*}[htbp]
   \centering
   \tabcolsep=0.35em
   \renewcommand{\arraystretch}{1.0}
   \caption{Estimation errors under different distortion. The distortion coefficient is set to $0.02$/$-0.02$ separately.}
   \label{tab:distortion_exp}
   \begin{threeparttable}
   \begin{tabular}{l | c | c c c c c c c c}
   \hline
      & $\delta\;[pix.]$ & $f_u\;[\%]$ & $f_v\;[\%]$ & $u_0\;[\%]$ & $v_0\;[\%]$ & $\bf R\;[^{\circ}]$ & ${\bf T}_{deg}\;[^{\circ}]$ & ${\bf T}_{scale}\;[\%]$ \\
      \hline
      $Bunny$ & 2.22/2.21 & $0.94/0.95$ & $0.93/0.95$ & $1.12/1.41$ & $1.09/1.88$ & $0.78/0.79$ & $0.83/0.81$ & $6.01/5.73$\\
      $Hood$ & 3.07/3.06 & $1.40/1.48$ & $1.84/1.87$ & $1.87/2.05$ & $1.91/1.78$ & $1.82/1.60$ & $1.41/1.47$ & $7.68/7.95$\\   
    \hline
   \end{tabular}
    \begin{tablenotes}
      \item \textmd{$\delta$ denotes the mean pixel offsets introduced by the lens distortion. }
    \end{tablenotes}
   \end{threeparttable}
\end{table*}

\begin{figure}[htb]
\centering 
\includegraphics[width=0.5\linewidth]{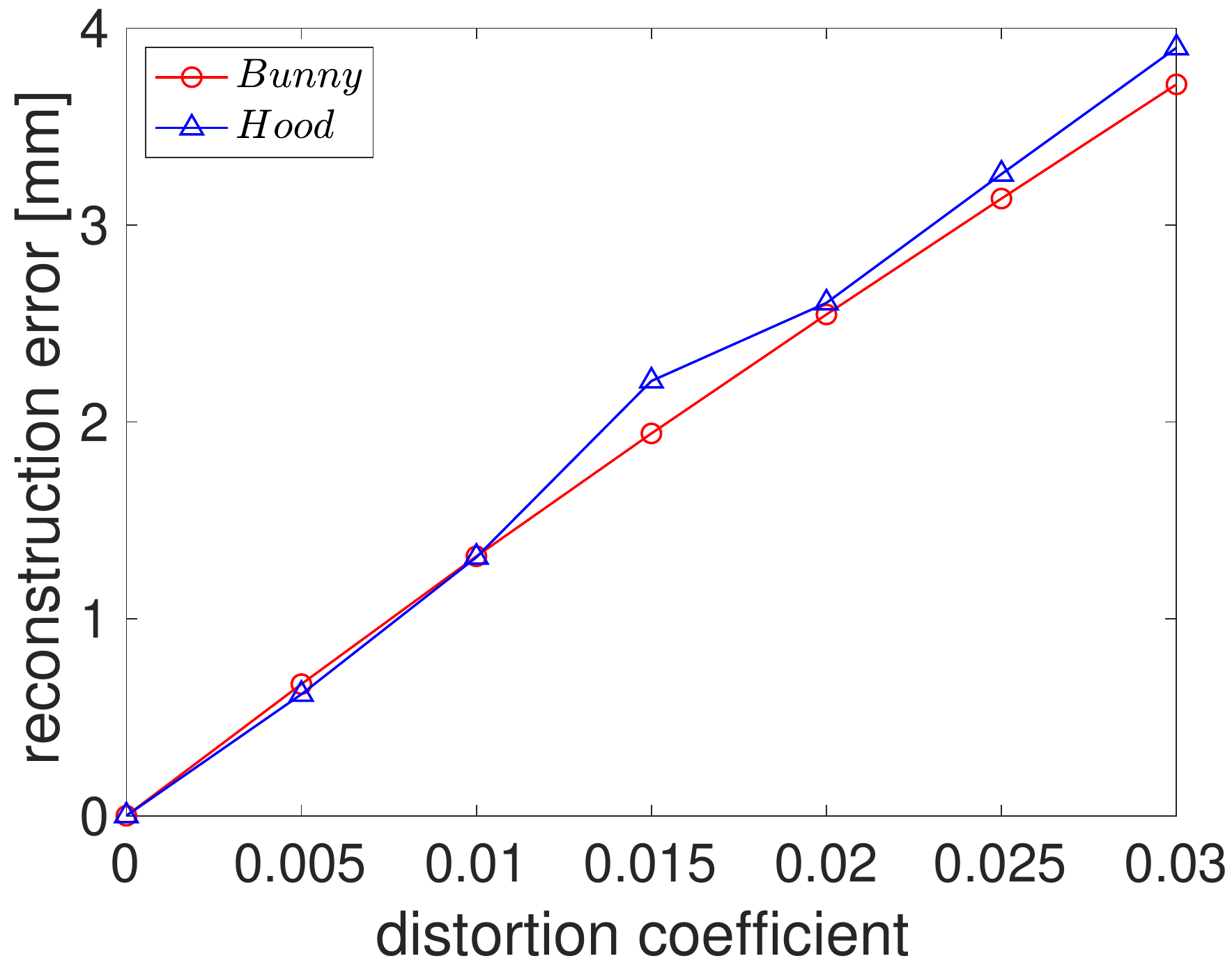}
\caption{Reconstruction results under different distortion coefficients.}
\label{fig:distortion}
\end{figure}

\begin{figure} [htbp]
   \centering
  \includegraphics[width=0.5\linewidth]{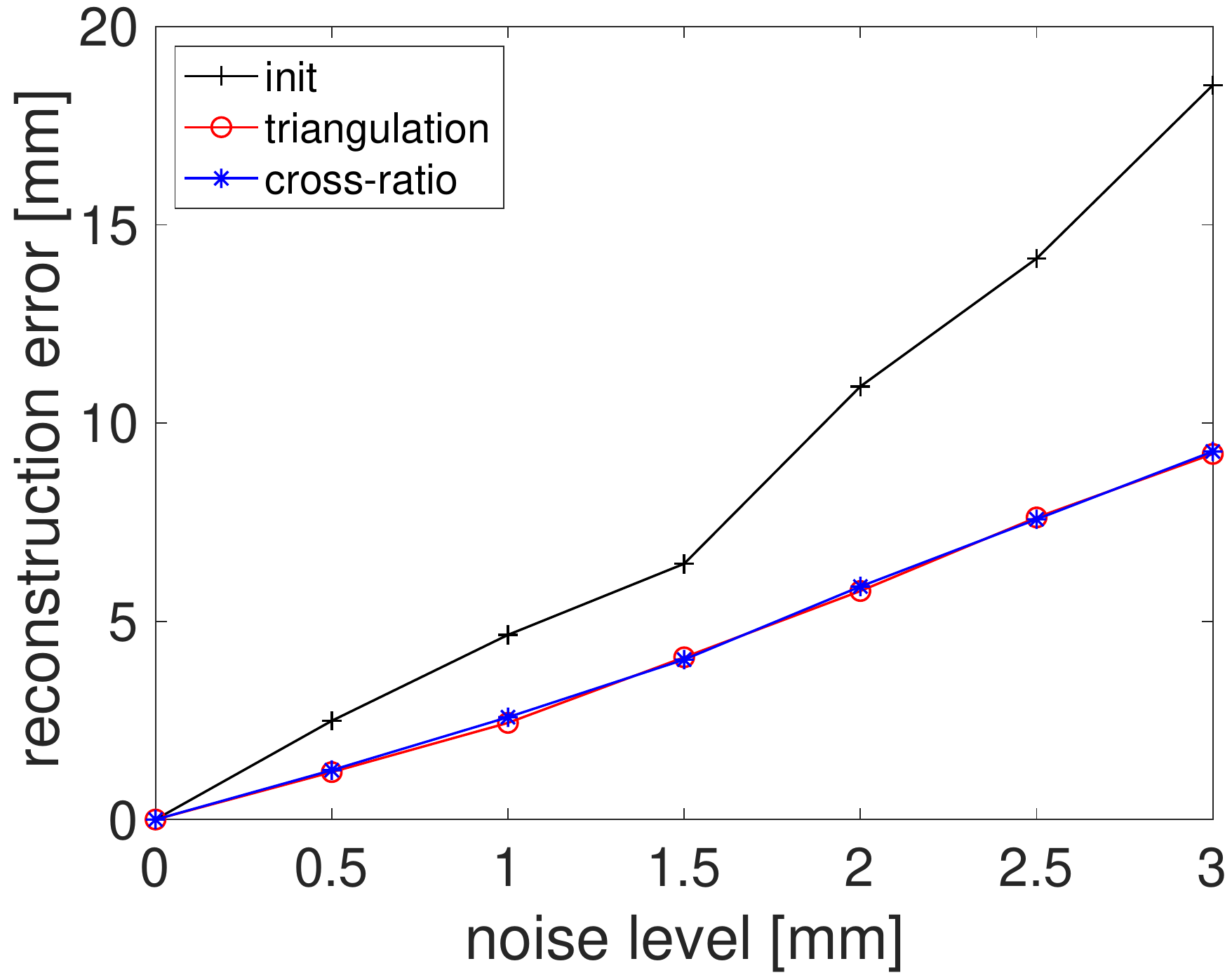}
   \caption{Surface reconstruction errors of \emph{bunny}. ``init'' indicates ray triangulation based on the initial camera projection matrix by our method in Section~\ref{initialization_E}; ``triangulation'' indicates ray triangulation based on the camera projection matrix by our cross-ratio based formulation; ``cross-ratio'' indicates our cross-ration based method.}
   \label{fig:triangulation_vs_cross}
\end{figure}

\begin{figure*}[htbp]
  \centering
   \tabcolsep=0.02cm
   \begin{tabular}{
         >{\centering\arraybackslash} m{0.190\textwidth}
         >{\centering\arraybackslash} m{0.190\textwidth}
         >{\centering\arraybackslash} m{0.190\textwidth}
         >{\centering\arraybackslash} m{0.190\textwidth}
         >{\centering\arraybackslash} m{0.190\textwidth}}
      \includegraphics[width=1\linewidth]{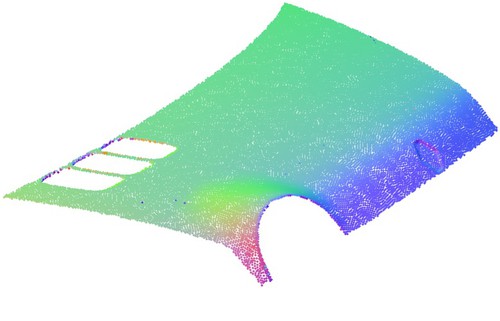} &
      \includegraphics[width=1\linewidth]{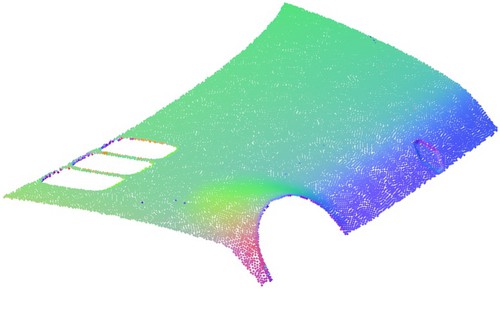} &
      \includegraphics[width=1\linewidth]{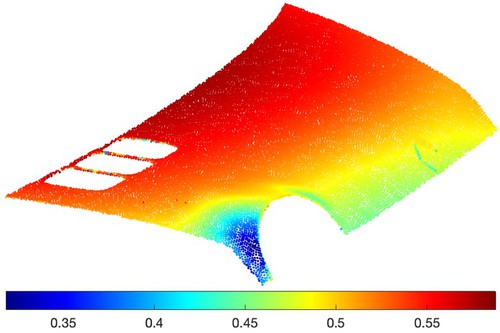} &
      \includegraphics[width=1\linewidth]{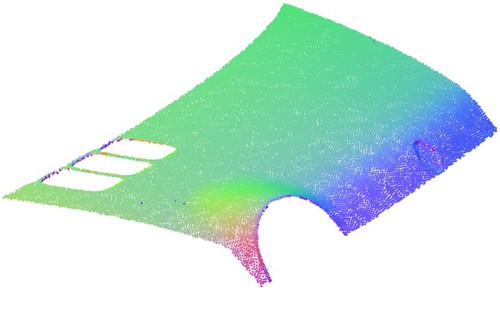} &
      \includegraphics[width=1\linewidth]{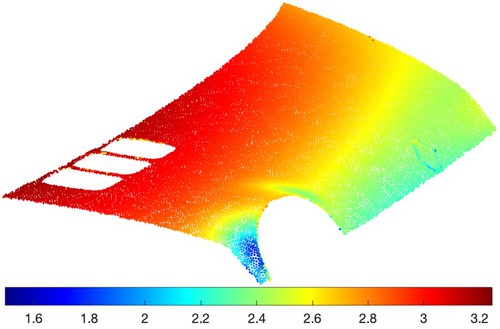} \\

      ground truth &
      $\sigma = 2.0$  &
      $\sigma = 2.0$ &
      $k_1 = 0.02$ &
      $k_1 = 0.02$\\
   \end{tabular}
   \caption{Normal reconstruction results of \emph{hood} shape. \textmd{Left to right: ground truth, normal estimation w/ correspondence noise $\sigma = 2.0$, normal estimation error w/ correspondence noise $\sigma = 2.0$, normal estimation w/ distortion coefficient $k_1 = 0.02$, normal estimation error w/ distortion coefficient $k_1 = 0.02$. The color in error maps indicates error in degree.}}
   \label{fig:hood_normal}
\end{figure*}

We consider the ideal pinhole camera model, therefore the potential lens distortion is not taken into account in our formulation. In practice, the lens may have radial and tangential distortion~\cite{Heikkila1997cvpr}. 
The one-parameter radial distortion is the most common form of distortion, which can be described by 
\begin{equation}
\begin{array}{c}
x_{2}=x_{1}(1+k_{1} r^{2}) \\
y_{2}=y_{1}(1+k_{1} r^{2})
\end{array},
\end{equation}
where $(x_1, y_1)$ and $(x_2, y_2)$ are the undistorted and distorted pixel locations in normalized image coordinates ($[-1, 1]$), $k_1$ is the radial distortion coefficient (positive/negative radial distortion is also known as \emph{barrel}/\emph{pincushion} distortion) of the lens and $r=\sqrt{x_{1}^{2}+y_{1}^{2}}$. 
To verify the robustness of our method to the lens distortion, we thoroughly evaluate the performance of our method on \emph{bunny} and \emph{hood} shapes by introducing radial distortion of different levels.
After introducing distortion, the pixel locations are denormalized to the original magnitude for camera pose estimation and surface reconstruction. 
We vary the distortion coefficient $k_1$ and measure the reconstruction errors on \emph{bunny} and \emph{hood} (see~\fref{fig:distortion}). In~\Tref{tab:distortion_exp}, we show the results of our approach under with $k_1$ of $0.02$/$-0.02$.  It can be seen that our method is generally robust to the lens distortion.

\begin{figure}[htb]
\centering 
\includegraphics[width=0.5\linewidth]{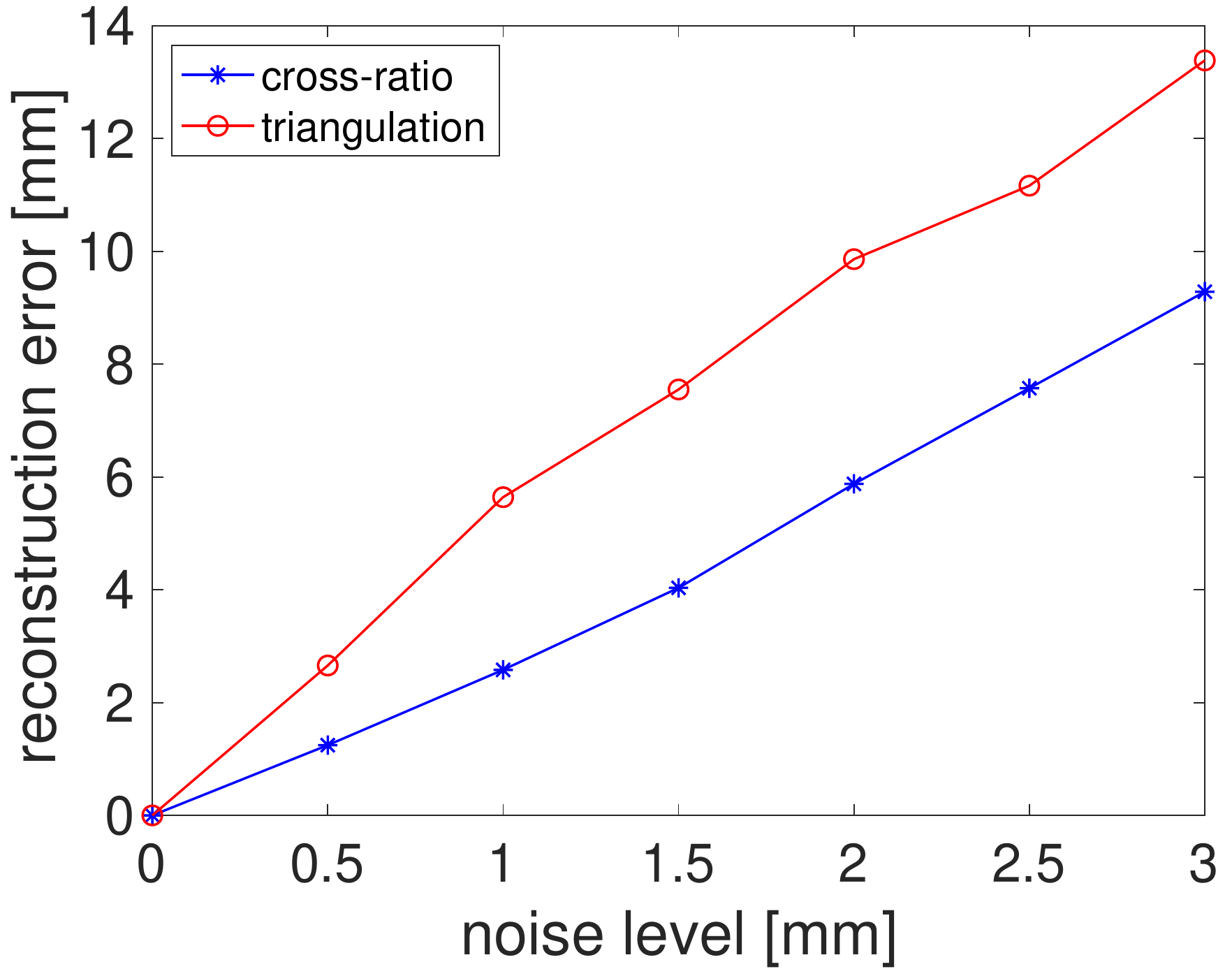}
\caption{Surface reconstruction errors of \emph{bunny} using cross-ratio and triangulation based optimization formulations respectively.}
\label{fig:opt_cross_tri}
\end{figure}

We further compare the surface reconstruction results between the ray triangulation and our cross-ratio based method in~\fref{fig:triangulation_vs_cross}, given the camera projection matrix estimated by our method. It can be seen that with the estimated camera projection matrix, the reconstruction accuracy of ray triangulation is on par with our cross-ratio based formulation. As the surface points can be obtained as the by-product of our cross-ratio based methods, the extra ray triangulation is no longer needed. Meanwhile, comparing with the ray triangulation results using our camera projection matrix initialization method in Section IV-C, our cross-ratio based method can significantly improve the reconstruction results.
In addition, we also explored the triangulation counterpart for the optimization problem in~(\ref{eq:non-linear}), by simply replacing the 3D point formulation in~(\ref{eq:M}) with ray triangulation (i.e., triangulating the incident and the reflected (visual) rays). The results are reported in~\Tref{tab:cross_tri} and~\fref{fig:opt_cross_tri}. Though the triangulation based formulation can also improve the results obtained by our constrained linear solution introduced in Section~\ref{initialization_E} (\Tref{tab:linear_p} vs~\Tref{tab:cross_tri}), the cross-ratio formulation performs notably better. Both formulations aim to minimize the point-to-point reprojection error. However, as our cross-ratio based formulation also enforces the cross-ratio constraint during optimization, it includes more geometric regularization than ray triangulation based formulation, leading to better performance.

\begin{table}[htbp]
   \centering
   \tabcolsep=0.35em
   \renewcommand{\arraystretch}{1.0}
   \caption{Comparison between ray-triangulation (TR) and cross-ratio (CR) formulations on \emph{bunny} under noise lv $\sigma = 2.0$ [$mm$].}
   \label{tab:cross_tri}
   \begin{threeparttable}
   \begin{tabular}{l | c c c c c c c c}
   \hline
      & $f_u\;[\%]$ & $f_v\;[\%]$ & $u_0\;[\%]$ & $v_0\;[\%]$ & $\bf R\;[^{\circ}]$ & ${\bf T}_{deg}\;[^{\circ}]$ & ${\bf T}_{scale}\;[\%]$ \\
      \hline
      $TR$ & $0.82$ & $0.82$ & $1.62$ & $1.49$ & $0.66$ & $0.46$ & $0.16$\\
      $CR$ & $0.11$ & $0.11$ & $0.18$ & $0.25$ & $0.08$ & $0.07$ & $0.13$\\   
    \hline
   \end{tabular}
   \end{threeparttable}
\end{table}

After obtaining the camera projection matrix and surface points by our cross-ratio based method, we can obtain the normal by simply calculating the bisector of the visual ray and incident ray. \Fref{fig:hood_normal} shows the normal estimation results on \emph{hood}. It can be seen that our method can estimate accurate normals. In this work, we focus more on recovering the point cloud because it contains the actual scale information of the object.

\subsubsection{Real Data Experiments}

\begin{figure}[htbp]
\centering 
  \includegraphics[width=0.6\linewidth]{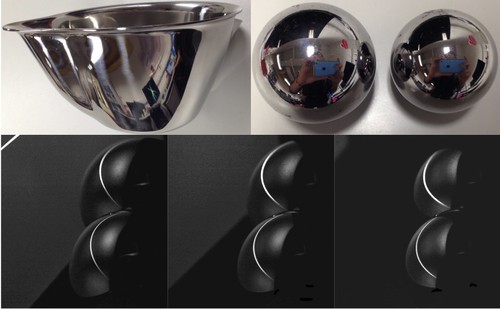}
\caption{\emph{Sauce boat} and \emph{two spheres} in real data experiments. \textmd{Top row: \emph{sauce boat} and \emph{two spheres} in real experiments. Bottom row: a sweeping line is reflected by \emph{two spheres} under three distinct positions of the LCD monitor while the camera and mirror surfaces are stationary.}}
\label{fig:realobjects}
\end{figure}

\begin{table*}[htbp]
   \centering
   \tabcolsep=4.5pt
   \renewcommand{\arraystretch}{1.0}
   \caption{Quantitative results of the real data experiments.}
   \label{tab:real_exp}
   \begin{threeparttable}
   \begin{tabular}{l | c c c c c c c c}
   \hline
      &  $f_u\;[pixel]$ & $f_v\;[pixel]$ & $u_0\;[pixel]$ & $v_0\;[pixel]$ & $\bf R\;[^{\circ}]$ & ${\bf T}_{deg}\;[^{\circ}]$ & ${\bf T}_{scale}\;[mm]$ & $S_{rms}\;[mm]$\\
      \hline
      $B_{uc}$ &  $36.70(0.63\%)$ & $21.99(0.38\%)$ & $99.10(5.03\%)$ & $100.00(8.13\%)$ & $9.12$ & $1.00$ & $19.16(8.23\%)$ & $2.55$\\
      $B_{cu}$ &  $-$ & $-$ & $-$ & $-$ & $8.63$ & $1.02$ & $16.15(6.96\%)$ & $2.29$\\   
      $B_{uu}$ &  $101.70(1.75\%)$ & $86.90(1.49\%)$ & $112.10(5.69\%)$ & $113.00(9.19\%)$ & $9.86$ & $1.99$ & $17.02(7.34\%)$ & $2.71$\\
      $S_{uc}$ &  $63.38(1.09\%)$ & $68.01(1.17\%)$ & $61.49(3.18\%)$ & $42.7(3.47\%)$ & $6.67$ & $1.78$ & $33.83(8.96\%)$ & $1.78$\\
      $S_{cu}$ &  $-$ & $-$ & $-$ & $-$ & $6.49$ & $1.61$ & $31.26(8.28\%)$ & $1.64$\\
      $S_{uu}$ &  $81.38(1.40\%)$ & $86.02(1.48\%)$ & $81.67(4.14\%)$ & $56.70(4.61\%)$ & $7.17$ & $2.13$ & $37.69(9.98\%)$ & $2.03$\\ 
    \hline
   \end{tabular}
    \begin{tablenotes}
      \item \textmd{$B$ and $S$ denote the results of \emph{sauce boat} and \emph{two spheres} respectively. The subscripts $uc$, $cu$, and $uu$ stand for experiments under (1) an uncalibrated camera and calibrated plane poses, (2) a calibrated camera and uncalibrated plane poses,  and (3) an uncalibrated camera and uncalibrated plane poses, respectively. The ground truth for the intrinsic parameters are $f_u = 5812.86$, $f_v = 5812.82$, and $(u_0, v_0) = (1971.95, 1230.02)$. $S_{rms}$ stands for the RMS reconstruction error.}
    \end{tablenotes}
   \end{threeparttable}
\end{table*}

\begin{figure*}[htbp]
   \centering
   \tabcolsep=0.1cm
   \renewcommand{\arraystretch}{0.5}
   \begin{tabular}{
         >{\centering\arraybackslash} m{0.235\textwidth}
         >{\centering\arraybackslash} m{0.235\textwidth}
         >{\centering\arraybackslash} m{0.235\textwidth}
         >{\centering\arraybackslash} m{0.235\textwidth}
}
      \includegraphics[width=\linewidth]{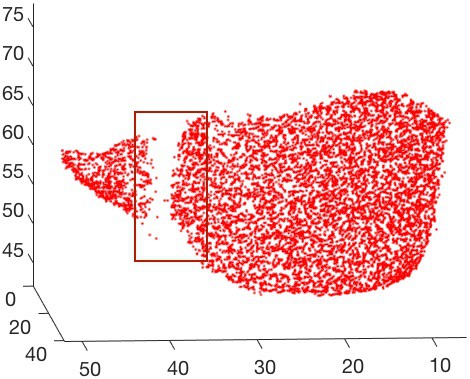} &
      \includegraphics[width=\linewidth]{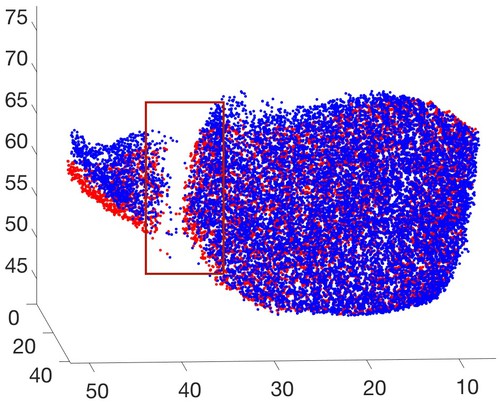} &
      \includegraphics[width=\linewidth]{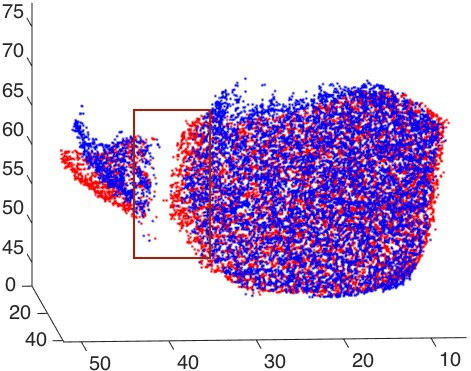} &
      \includegraphics[width=\linewidth]{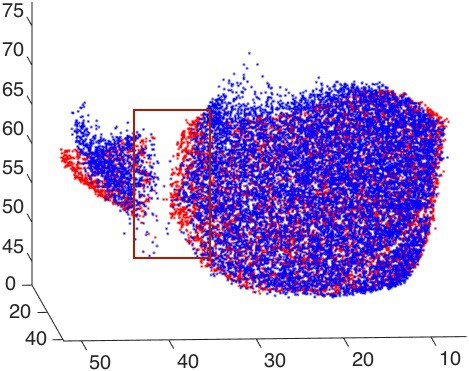} \\
      \includegraphics[width=\linewidth]{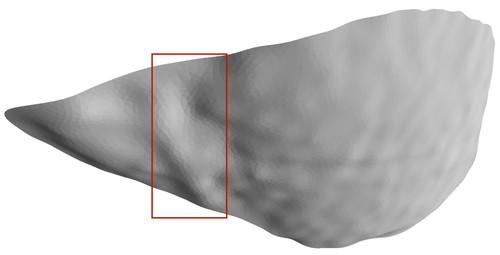} &
      \includegraphics[width=\linewidth]{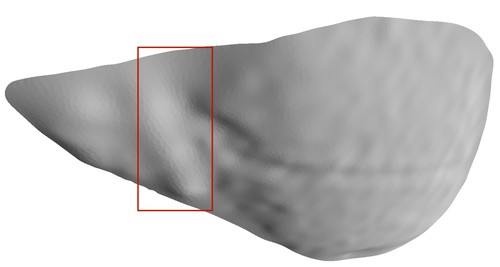} &
      \includegraphics[width=\linewidth]{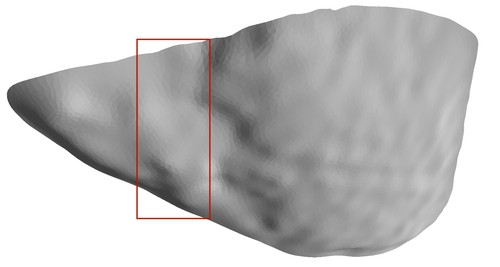} &
      \includegraphics[width=\linewidth]{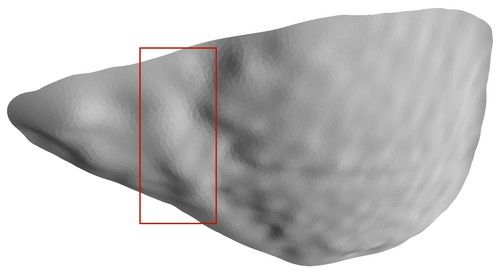} \\
      (a) &
      (b) &
      (c) &
      (d) \\
      \includegraphics[width=\linewidth]{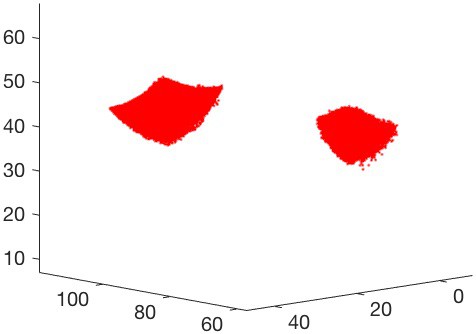} &
      \includegraphics[width=\linewidth]{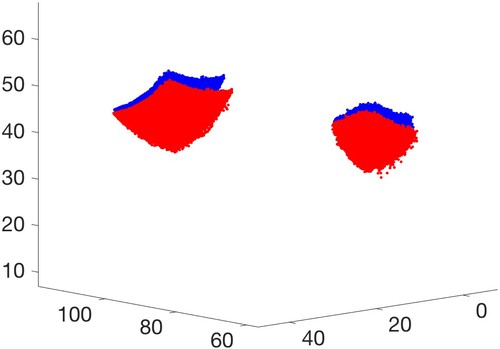} &
      \includegraphics[width=\linewidth]{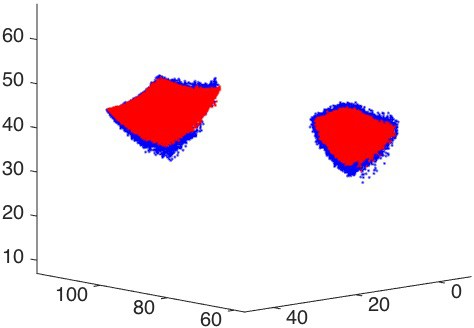} &
      \includegraphics[width=\linewidth]{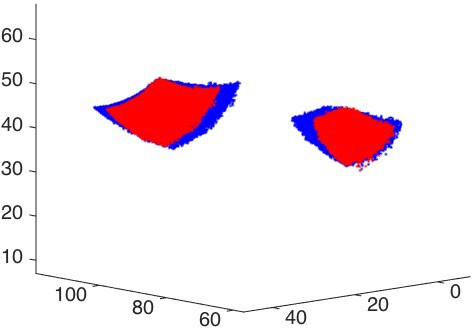} \\
      \includegraphics[width=\linewidth]{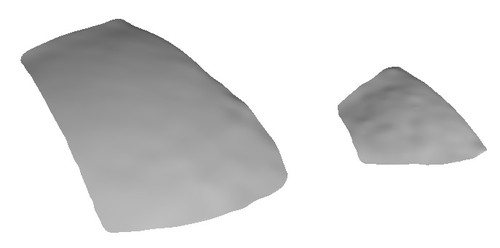} &
      \includegraphics[width=\linewidth]{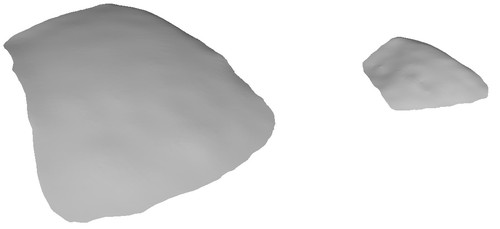} &
      \includegraphics[width=\linewidth]{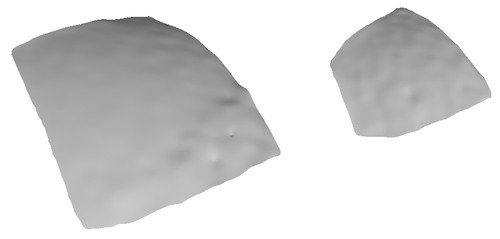} &
      \includegraphics[width=\linewidth]{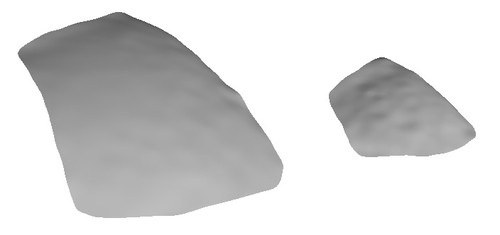} \\
      (e) &
      (f) &
      (g) &
      (h) \\
   \end{tabular}
   \caption{Reconstruction results of \emph{sauce boat} and \emph{two spheres}. \textmd{(a)-(d): reconstructions of \emph{sauce boat}. Results are obtained under (a) calibrated camera with calibrated plane poses (this result is treated as ground truth and overlaid in (b), (c) and (d) for comparison (red)); (b) uncalibrated camera with calibrated plane poses (blue); (c) calibrated camera with uncalibrated plane poses (blue); (d) uncalibrated camera with uncalibrated plane poses (ours, blue). Note the missing regions (in red rectangle) in the reconstructed point clouds are filled by the mesh generation algorithm and should be ignored in comparing the surface meshes. (e)-(h): reconstructions of \emph{two spheres}.}}
   \label{fig:results_real}
   \end{figure*}

We evaluated our method on a \emph{sauce boat} and \emph{two spheres}
respectively (see Fig.~\ref{fig:realobjects}). We captured images using a $Canon$ $EOS$ $40D$ digital camera with a $24$-$70$ $mm$ lens. A $19$ $inch$ LCD monitor was used as a reference plane and was placed at three different locations. We follow \cite{Kutulakos2008ijcv, han2015cvpr, han2018ijcv} to display sweeping stripes on the monitor to establish reflection correspondences. In particular, at each location, the LCD monitor shows a thin stripe sweeping across the screen in vertical direction and then in horizontal direction. For each position of the sweeping stripe, we capture an image. We then obtain a sequence of images w.r.t the vertically sweeping stripe and a sequence of images w.r.t the horizontally sweeping stripe. We can then obtain the correspondence on the reference plane for each image point by examining its corresponding two sequences of intensity values. The position of the stripe that produces the peak intensity value in each sweeping direction then gives the position of the correspondence on the reference plane. In order to improve the accuracy of the peak localization, we fit a quadratic curve to the intensity profile in the neighborhood of the sampled peak value, and solve for the exact peak analytically.

After establishing reflection correspondences, we first estimated the relative poses of the reference plane using our method described in Section~\ref{sec:planepose}. We then formed 3D lines from the reflection correspondences on the reference plane under the two poses that were furthest apart (e.g., $P_0$ and $P_2$ in Fig.~\ref{fig:cross_ratio}). These 3D lines were used to obtain a preliminary solution of projection matrix using the linear method in Section~\ref{closeformsolution}, which was then used to initialize the nonlinear optimization described in Section~\ref{robustestimation}.

To evaluate our method, we calibrated the camera and reference plane poses using the \emph{Camera Calibration Toolbox for Matlab} \cite{Bouguet_MatlabCalibrationToolBox}. We used the calibration result to estimate the surface and treated it as the ground truth. This result was compared against the result obtained using 1) uncalibrated camera and calibrated plane poses, 2) calibrated camera and uncalibrated plane poses, and 3) uncalibrated camera and uncalibrated plane poses. Fig.~\ref{fig:results_real} shows the reconstructed surfaces and Table~\ref{tab:real_exp} shows the numerical errors. We aligned each estimated surface with the ground truth by a rigid body transformation using the method in \cite{besl1992pami} before computing the reconstruction error. The RMS reconstruction errors are below $3$ $mm$. $f_u$ and $f_v$ errors are below ${2\%}$. $u_0$, $v_0$ and ${\bf T}_{scale}$ errors are below ${10\%}$. The angular errors are below $10^\circ$ for ${\bf R}$  and below $3^\circ$ for ${\bf T}$. The errors in intrinsics and extrinsics are larger than those in the synthetic experiments. This is reasonable since accurate reflection correspondences in real cases are difficult to obtain due to the large and complex distortion caused by the mirror surface and varying lighting condition. Besides, degradation from optics, such as optics aberrations, will also introduce errors for establishing reflections correspondences.
The quality of the reflection correspondences is also a key factor for other existing mirror surface reconstruction methods. Note that existing methods are designed under certain assumptions (e.g., convex, $C^n$ continuity, etc),  and their setups are carefully tailored or require special equipments. Besides, there is no publicly available dataset that can serve as a benchmark for existing methods. As a result, it is challenging to make a fair comparison with existing methods on real data. Therefore, we do not include comparison with other existing methods on real data.

\section{Conclusions}
\label{conclusion}
A novel method is introduced for mirror surface reconstruction. Our method works under an uncalibrated setup and can recover the camera intrinsics and extrinsics, along with the surface. We proposed an analytical solution for the reference plane relative pose estimation, a closed-form solution for camera projection matrix estimation, and a cross-ratio based formulation to achieve a robust estimation of both camera projection matrix and the mirror surface. The proposed method only needs reflection correspondences as input and removes the restrictive assumptions of known motions, $C^n$ continuity of the surface, and calibrated camera(s) that are being used by other existing methods. This greatly simplifies the challenging problem of mirror surface recovery. We believe our work can provide a meaningful insight towards solving this problem. 

Our method does have a few limitations. We assume the surface is perfectly reflective and our formulation depends on the law of reflection. Thus our method may not work for other mirror-like surfaces that do not follow the law of reflection, or mirror surfaces that do not reflect the full energy of the incident rays. Although our cross-ratio based formulation does not encounter degenerate cases, degenerate cases may occur in the relative pose estimation of the reference plane. Efforts are needed to explore methods without degeneracy for estimating the relative poses of the reference plane from reflection correspondences. Meanwhile, the region that can be reconstructed by our method depends on the size of the reference plane. For mirror surfaces with complex normal distribution, the reference plane may need to be carefully placed at different locations facing different regions of the mirror surface, and the reference plane should be  moved three times at each location. Despite an increased number of images, extra efforts are also required to fuse these reconstructed regions together. 
Besides, in our formulation, we did not consider inter-reflection, which is likely to happen for regions with relatively sharp changes in the mirror surface. In the future, we would like to develop methods to recover complete surfaces and further investigate inter-reflection cases.

\section*{Acknowledgement}
This project is supported by a grant from the Research Grant Council of the Hong Kong (SAR), China, under the project HKU 17203119 and a grant (DE180100628) from Australian Research Council (ARC).

\ifCLASSOPTIONcaptionsoff
  \newpage
\fi

\bibliographystyle{IEEEtran}
\bibliography{reference_mirror}

\begin{thebibliography}{10}
\providecommand{\url}[1]{#1}
\csname url@samestyle\endcsname
\providecommand{\newblock}{\relax}
\providecommand{\bibinfo}[2]{#2}
\providecommand{\BIBentrySTDinterwordspacing}{\spaceskip=0pt\relax}
\providecommand{\BIBentryALTinterwordstretchfactor}{4}
\providecommand{\BIBentryALTinterwordspacing}{\spaceskip=\fontdimen2\font plus
\BIBentryALTinterwordstretchfactor\fontdimen3\font minus
  \fontdimen4\font\relax}
\providecommand{\BIBforeignlanguage}[2]{{%
\expandafter\ifx\csname l@#1\endcsname\relax
\typeout{** WARNING: IEEEtran.bst: No hyphenation pattern has been}%
\typeout{** loaded for the language `#1'. Using the pattern for}%
\typeout{** the default language instead.}%
\else
\language=\csname l@#1\endcsname
\fi
#2}}
\providecommand{\BIBdecl}{\relax}
\BIBdecl

\bibitem{han2016cvpr}
K.~Han, K.-Y.~K. Wong, D.~Schnieders, and M.~Liu, ``Mirror surface
  reconstruction under an uncalibrated camera,'' in \emph{CVPR}, 2016, pp.
  1772--1780.

\bibitem{Balzer2010measurement}
J.~Balzer and S.~Werling, ``Principles of shape from specular reflection,''
  \emph{Measurement}, vol.~43, no.~10, pp. 1305--1317, 2010.

\bibitem{ihrke2010cgf}
I.~Ihrke, K.~N. Kutulakos, H.~P.~A. Lensch, M.~Magnor, and W.~Heidrich,
  ``Transparent and specular object reconstruction,'' \emph{Computer Graphics
  Forum}, vol.~29, pp. 2400--2426, 2010.

\bibitem{reshetouski2013}
I.~Reshetouski and I.~Ihrke, ``Mirrors in computer graphics, computer vision
  and time-of-flight imaging,'' \emph{Lect. Notes Comput. Sc.}, vol. 8200, pp.
  77--104, 2013.

\bibitem{oren1996ijcv}
M.~Oren and S.~K. Nayar, ``A theory of specular surface geometry,''
  \emph{IJCV}, vol.~24, no.~2, pp. 105--124, 1996.

\bibitem{roth2006cvpr}
S.~Roth and M.~J. Black, ``Specular flow and the recovery of surface
  structure,'' in \emph{CVPR}, 2006, pp. 1869--1876.

\bibitem{adato2007iccv}
Y.~Adato, Y.~Vasilyev, O.~Ben-Shahar, and T.~Zickler, ``Toward a theory of
  shape from specular flow,'' in \emph{ICCV}, 2007, pp. 1--8.

\bibitem{adato2010pami}
Y.~Adato, Y.~Vasilyev, T.~Zickler, and O.~Ben-Shahar, ``Shape from specular
  flow,'' \emph{PAMI}, vol.~32, no.~11, pp. 2054--2070, 2010.

\bibitem{Vasilyev2011cvpr}
Y.~Vasilyev, T.~Zickler, S.~Gortler, and O.~Ben-Shahar, ``Shape from specular
  flow: Is one flow enough?'' in \emph{CVPR}, 2011, pp. 2561--2568.

\bibitem{canas2009iccv}
G.~D. Canas, Y.~Vasilyev, Y.~Adato, T.~Zickler, S.~Gortler, and O.~Ben-Shahar,
  ``A linear formulation of shape from specular flow,'' in \emph{ICCV}, 2009,
  pp. 191--198.

\bibitem{sankaranarayanan2010cvpr}
A.~Sankaranarayanan, A.~Veeraraghavan, O.~Tuzel, and A.~Agrawal, ``Specular
  surface reconstruction from sparse reflection correspondences,'' in
  \emph{CVPR}, 2010, pp. 1245--1252.

\bibitem{Savarese2001cvpr}
S.~Savarese and P.~Perona, ``Local analysis for 3d reconstruction of specular
  surfaces.'' in \emph{CVPR}, 2001, pp. 738--745.

\bibitem{Savarese2002eccv}
S.~Savarese and P.~Perona, ``Local analysis for 3d reconstruction of specular
  surfaces -- part ii,'' in \emph{ECCV}, 2002, pp. 759--774.

\bibitem{Rozenfeld2011pami}
S.~Rozenfeld, I.~Shimshoni, and M.~Lindenbaum, ``Dense mirroring surface
  recovery from 1d homographies and sparse correspondences,'' \emph{PAMI},
  vol.~33, no.~2, pp. 325--337, 2011.

\bibitem{Liu2015pami}
M.~Liu, R.~Hartley, and M.~Salzmann, ``Mirror surface reconstruction from a
  single image,'' \emph{PAMI}, vol.~37, no.~4, pp. 760--773, 2015.

\bibitem{bonfort2003iccv}
T.~Bonfort and P.~Sturm, ``Voxel carving for specular surfaces,'' in
  \emph{ICCV}, 2003, pp. 691--696.

\bibitem{bonfort2006accv}
T.~Bonfort, P.~Sturm, and P.~Gargallo, ``General specular surface
  triangulation,'' in \emph{ACCV}, 2006, pp. 872--881.

\bibitem{Sturm2006accv}
P.~Sturm and T.~Bonfort, ``How to compute the pose of an object without a
  direct view,'' in \emph{ACCV}, 2006, pp. 21--31.

\bibitem{Nehab2008cvpr}
D.~Nehab, T.~Weyrich, and S.~Rusinkiewicz, ``Dense 3d reconstruction from
  specularity consistency,'' in \emph{CVPR}, 2008, pp. 1--8.

\bibitem{Weinmann2013iccv}
M.~Weinmann, A.~Osep, R.~Ruiters, and R.~Klein, ``Multi-view normal field
  integration for 3d reconstruction of mirroring objects.'' in \emph{ICCV},
  2013, pp. 2504--2511.

\bibitem{Balzeretal2014}
J.~Balzer, D.~Acevedo-Feliz, S.~Soatto, S.~H{\"o}fer, M.~Hadwiger, and
  J.~Beyerer, ``Cavlectometry: Towards holistic reconstruction of large mirror
  objects,'' in \emph{International Conference on 3D Vision (3DV)}, 2014, pp.
  448--455.

\bibitem{Tin2016iccp}
S.-K. Tin, J.~Ye, M.~Nezamabadi, and C.~Chen, ``3d reconstruction of
  mirror-type objects using efficient ray coding,'' in \emph{ICCP}, 2016.

\bibitem{lu2019iccp}
J.~Lu, Y.~Ji, J.~Yu, and J.~Ye, ``Mirror surface reconstruction using
  polarization field,'' in \emph{ICCP}, 2019, pp. 1--9.

\bibitem{Kutulakos2008ijcv}
K.~N. Kutulakos and E.~Steger, ``A theory of refractive and specular {3D} shape
  by light-path triangulation,'' \emph{IJCV}, vol.~76, pp. 13--29, 2008.

\bibitem{Liu2010accv}
M.~Liu, K.-Y.~K. Wong, Z.~Dai, and Z.~Chen, ``Specular surface recovery from
  reflections of a planar pattern undergoing an unknown pure translation,'' in
  \emph{ACCV}, vol.~2, 2010, pp. 647--658.

\bibitem{Perdigoto2013cviu}
L.~Perdigoto and H.~Araujo, ``Calibration of mirror position and extrinsic
  parameters in axial non-central catadioptric systems,'' \emph{CVIU}, vol.
  117, pp. 909--921, 2013.

\bibitem{ramalingam2005cvpr}
S.~Ramalingam, P.~Sturm, and S.~K. Lodha, ``Towards complete generic camera
  calibration,'' in \emph{CVPR}, 2005, pp. 1093--1098.

\bibitem{Ramalingam_2015_CVPR}
S.~Ramalingam, M.~Antunes, D.~Snow, G.~Hee~Lee, and S.~Pillai, ``Line-sweep:
  Cross-ratio for wide-baseline matching and 3d reconstruction,'' in
  \emph{CVPR}, 2015, pp. 1238--1246.

\bibitem{Zhang1998techreport}
Z.~Zhang, ``A flexible new technique for camera calibration,'' \emph{Technical
  Report, MSR-TR-98-71}, pp. 1--22, 1998.

\bibitem{Baker1999ijcv}
S.~Baker and S.~Nayar, ``A theory of single-viewpoint catadioptric image
  formation,'' \emph{IJCV}, vol.~35, no.~2, pp. 175--196, 1999.

\bibitem{hartleyMVG}
R.~Hartley and A.~Zisserman, \emph{Multiple View Geometry in Computer Vision},
  2nd~ed.\hskip 1em plus 0.5em minus 0.4em\relax Cambridge University Press,
  2004.

\bibitem{kazhdan2013tog}
M.~Kazhdan and H.~Hoppe, ``Screened {P}oisson surface reconstruction,''
  \emph{ACM Transactions on Graphics (TOG)}, vol.~32, pp. 29:1--29:13, 2013.

\bibitem{Heikkila1997cvpr}
J.~Heikkila and O.~Silven, ``A four-step camera calibration procedure with
  implicit image correction,'' in \emph{CVPR}, 1997, pp. 1106--1112.

\bibitem{han2015cvpr}
K.~Han, K.-Y.~K. Wong, and M.~Liu, ``A fixed viewpoint approach for dense
  reconstruction of transparent objects,'' in \emph{CVPR}, 2015, pp.
  4001--4008.

\bibitem{han2018ijcv}
K.~Han, K.-Y.~K. Wong, and M.~Liu, ``Dense reconstruction of transparent
  objects by altering incident light paths through refraction,'' \emph{IJCV},
  vol. 126, no.~5, pp. 460--475, 2018.

\bibitem{Bouguet_MatlabCalibrationToolBox}
J.-Y. Bouguet, ``Camera calibration toolbox for matlab,''
  http://www.vision.caltech.edu/bouguetj/calib\_doc/, 2008.

\bibitem{besl1992pami}
P.~J. Besl and H.~D. McKay, ``A method for registration of 3-d shapes,''
  \emph{PAMI}, vol.~14, no.~2, pp. 239--256, 1992.

\end{thebibliography}


\begin{thebibliography}{10}
\end{thebibliography}

\end{document}


\title{Fixed Viewpoint Mirror Surface Reconstruction under an Uncalibrated Camera \\ -Supplementary Material-}

\author{\quad Kai Han$^1$ \quad Miaomiao Liu$^2$ \quad Dirk Schnieders$^3$ \quad Kwan-Yee K. Wong$^3$  \vspace{0.3em}\\
\quad$^1$University of Bristol  \quad$^2$The Australian National University \quad $^3$The University of Hong Kong \\
}
\maketitle

\section{Pl\"{u}cker Coordinates} 
A 3D line can be described by a Pl\"{u}cker matrix $\bf {L} =  \mathbf{QP^T - PQ^T} =$
\begin{equation}
      \begin{bmatrix}
         0 & q_1p_2-q_2p_1 & q_1p_3-q_3p_1 & q_1p_4-q_4p_1\\
         q_2p_1-q_1p_2 & 0 & q_2p_3-q_3p_2 & q_2p_4-q_4p_2\\
         q_3p_1-q_1p_3 & q_3p_2-q_2p_3 & 0 & q_3p_4-q_4p_3\\
         q_4p_1-q_1p_4 & q_4p_2-q_2p_4 & q_4p_3-q_3p_4 & 0
      \end{bmatrix},
\end{equation}
where ${\bf P} = [p_1\ p_2\ p_3\ p_4]^{\rm T}$ and ${\bf Q} = [q_1\ q_2\ q_3\ q_4]^{\rm T}$ are the homogeneous representations of two distinct 3D points on the line. 
Since ${\bf L}$ is skew-symmetric, it can be represented simply by a Pl\"{u}cker vector $\mathcal{L}$ consisting of its $6$ distinct non-zero elements
\begin{equation}
  \rm
  \mathcal{L}
  = 
  \begin{bmatrix}
  l_1\\
  l_2\\ 
  l_3\\ 
  l_4\\ 
  l_5\\ 
  l_6
  \end{bmatrix}
  = 
  \begin{bmatrix}
  q_1p_2-q_2p_1\\
  q_1p_3-q_3p_1\\ 
  q_1p_4-q_4p_1\\ 
  q_2p_3-q_3p_2\\ 
  q_3p_4-q_4p_3\\ 
  q_4p_2-q_2p_4
  \end{bmatrix}.
  \label{eq:L}
\end{equation}

Dually, a matrix ${\bf \bar { L}}$ can be constructed from two distinct planes with homogeneous representations $\bf \hat P$ and $\bf \hat Q$ as $\rm {\bf{\bar L} = {\hat Q}}{\bf \hat P}^T-{\bf{\hat P}}{\bf \hat Q}^T$. The {\em dual Pl\"{u}cker vector} can be constructed directly from $\bf \bar L$ or by rearranging the elements of $\mathcal{L}$ as
\begin{equation}
\mathcal{\bar L} = [l_5\ l_6\ l_4\ l_3\ l_1\ l_2]^{\rm T}.\label{eq:dualPluck}  
\end{equation}

Let ${\bf A} = [a_1\ a_2\ a_3]^{\rm T}$  and ${\bf B} = [b_1\ b_2\ b_3]^{\rm T}$ be two distinct 3D points in Cartesian coordinates. Geometrically, the line defined by these points can be represented by a direction vector $\boldsymbol{\omega} = {\bf (A - B)} = [l_3\ -l_6\ l_5]^{\rm T}$ and a moment vector $\boldsymbol{\nu} = {\bf (A \times B)} = [l_4\ -l_2\ l_1]^{\rm T}$, which define the line up to a scalar factor. 

Two 3D lines $\mathcal{L}$ and $\mathcal{L}^\prime$ can either be skew or coplanar. The geometric requirement for the latter case is that the dot product between the direction vector of the first line and the moment vector of the second line should equal the negative of the dot product between the direction vector of the second line and the moment vector of the first line. Let the two lines have direction vectors $\boldsymbol \omega$, $\boldsymbol{\omega}^\prime$ and moment vectors $\boldsymbol \nu$, $\boldsymbol{\nu}^\prime$, respectively. They are coplanar (i.e., either coincident or intersect) if and only if 
\begin{equation}
  \boldsymbol{\omega}\cdot \boldsymbol{\nu}^\prime + \boldsymbol{\nu}\cdot \boldsymbol{\omega}^\prime = 0 \Leftrightarrow
  {\mathcal{L}\cdot \mathcal{\bar L}^{\prime}} = 0. \label{eq:pcond1}
\end{equation}

Note that a Pl\"{u}cker vector is not any arbitrary $6$-vector. A valid Pl\"{u}cker vector must always intersect itself, i.e.,
\begin{equation}
  \mathcal{L}\cdot \mathcal{\bar L} = 0 \Leftrightarrow det({\bf L}) = 0 .
  \label{eq:pcond2}
\end{equation}

\section{Camera projection matrix vs. line projection matrix}
Here we show the details for the conversion between camera projection matrix and the line projection matrix.
First, consider the case of transforming a point projection matrix to its equivalent line projection matrix. Referring to Fig. 3 in our main paper, the plane ${\bf P}_{1*}{\bf X} = {\bf 0}$ is defined by the camera center and the line $u = 0$ in the image plane. Similarly, ${\bf P}_{2*}{\bf X} = {\bf 0}$ is defined by the camera center and the line $v = 0$ in the image plane. Finally, the plane equation ${\bf P}_{3*}{\bf X} = {\bf 0}$ holds for all points with pixel coordinates $s = 0$. We can obtain the $i$-th row of $\mathcal P$ by the intersection of rows $j$ and $k$ of $\bf P$, i.e.,
\begin{equation}
     \mathcal {P}_{i*}^{\rm T} 
     =
     \begin{bmatrix}
        \rho_{i1}\\
        \rho_{i2}\\
        \rho_{i3}\\
        \rho_{i4}\\
        \rho_{i5}\\
        \rho_{i6}
     \end{bmatrix}
     = (-1)^{(i+1)}
     \begin{bmatrix}
        p_{j3}p_{k4} - p_{j4}p_{k3}\\
        p_{j4}p_{k2} - p_{j2}p_{k4}\\
        p_{j2}p_{k3} - p_{j3}p_{k2}\\
        p_{j1}p_{k4} - p_{j4}p_{k1}\\
        p_{j1}p_{k2} - p_{j2}p_{k1}\\
        p_{j1}p_{k3} - p_{j3}p_{k1}
     \end{bmatrix} ,
  \end{equation}
where $\mathbf P_{i*}$ is the $i$-th row of $\mathbf P$, $\mathcal P_{i*} $ is the $i$-th row of $\mathcal P$, $i \neq j \neq k \in \{1, 2, 3\}$ and $j < k$. Note that ${\mathcal P}_{i*} ^{\rm T}$ is the dual Pl\"{u}cker vector of $(-1)^{i+1}({\bf P}_{j*}^{\rm T} \wedge{\bf P}_{k*}^{\rm T} )$, i.e., the intersection of the $j$-th with the $k$-th row of $\bf P$. The sign here controls the order of intersection, i.e., $({\bf P}_{j*}^{\rm T} \wedge{\bf P}_{k*}^{\rm T} ) = -({\bf P}_{k*}^{\rm T} \wedge{\bf P}_{j*}^{\rm T} )$. Dually, we can obtain the $i$-th row of $\bf P$ by the intersection of rows $j$ and $k$ of $\mathcal P$, which results in the homogeneous plane
\begin{equation}
\begin{split}
\mathbf{P}_{i*}^{\rm T}& =(-1)^{(i+1)} 
     \begin{bmatrix}
     \boldsymbol{\omega}_j\times\boldsymbol{\omega}_k\\
     \boldsymbol{\nu}_j\cdot\boldsymbol{\omega}_k
     \end{bmatrix}
\\
&        = (-1)^{(i+1)}
     \begin{bmatrix}
        \rho_{j5}\rho_{k6} - \rho_{j6}\rho_{k5}\\
        \rho_{j5}\rho_{k3} - \rho_{j3}\rho_{k5}\\
        \rho_{j6}\rho_{k3} - \rho_{j3}\rho_{k6}\\
        \rho_{j4}\rho_{k3} + \rho_{j2}\rho_{k6} + \rho_{j1}\rho_{k5}
     \end{bmatrix} 
     \\
\end{split} ,
\end{equation}
where again $i \neq j \neq k \in \{1, 2, 3\}$ with $j < k$, $\boldsymbol{\omega}_j$ is the direction vector of ${\mathcal P}_{j*}^{\rm T}$ and $\boldsymbol{\nu}_j$ is the moment vector of ${\mathcal P}_{j*}^{\rm T}$.

\section{Plane Pose Estimation from Reflection Correspondences}

As in Section III,%
we can obtain the solution space spanned by two vector basis, $\mathbf{d}_{1}$ and $\mathbf{d}_{2}$. $\mathbf{W}$ is then parameterized as 
\begin{equation}
\mathbf{W}=\alpha\left(\mathbf{d}_{1}+\beta \mathbf{d}_{2}\right), 
\label{eq:w_sol}
\end{equation}
where 
\begin{gather}
\mathbf{d}_{1}=\left(d_{1}^{1},\ d_{1}^{2},\ d_{1}^{3},\ \cdots,\ d_{1}^{24}\right),\\
\mathbf{d}_{2}=\left(d_{2}^{1},\ d_{2}^{2},\ d_{2}^{3},\ \cdots,\ d_{2}^{24}\right).
\end{gather}
The definition of $\mathbf{W}$, $\mathcal{A}$ and $\mathcal{B}$ are recollected in detail as follows
\begin{gather}
\mathbf{W}=\left[\mathcal{A}_{1 *}\ \mathcal{A}_{2 *}\ \mathcal{A}_{3 *}\ \mathcal{B}_{1 *}\ \mathcal{B}_{2 *}\ \mathcal{B}_{3 *}\ \mathcal{N}_{3 *}\ \mathcal{M}_{3 *}\right]^{\mathrm{T}}\label{eq:w},\\
\mathcal{A}=\mathcal{N}_{3*}^{\mathrm{T}} \mathcal{M}_{1 *}-\mathcal{N}_{1* }^{\mathrm{T}} \mathcal{M}_{3 *},\\
\mathcal{B}=\mathcal{N}_{3*}^{\mathrm{T}} \mathcal{M}_{2 *}-\mathcal{N}_{2* }^{\mathrm{T}} \mathcal{M}_{3 *}.
\end{gather}
According to the element-wise equality between~\Eref{eq:w_sol} and~\Eref{eq:w}, we have
\begin{eqnarray}
A_{11}=n_{31} m_{11}-n_{11} m_{31}=\alpha\left(d_{1}^{1}+\beta d_{2}^{1}\right),\label{eq:a11}\\
A_{12}=n_{31} m_{12}-n_{11} m_{32}=\alpha\left(d_{1}^{2}+\beta d_{2}^{2}\right),\label{eq:a12}\\
A_{13}=n_{31} m_{13}-n_{11} m_{33}=\alpha\left(d_{1}^{3}+\beta d_{2}^{3}\right),\label{eq:a13}\\
A_{21}=n_{32} m_{11}-n_{12} m_{31}=\alpha\left(d_{1}^{4}+\beta d_{2}^{4}\right),\label{eq:a21}\\
A_{22}=n_{32} m_{12}-n_{12} m_{32}=\alpha\left(d_{1}^{5}+\beta d_{2}^{5}\right),\label{eq:a22}\\
A_{23}=n_{32} m_{13}-n_{12} m_{33}=\alpha\left(d_{1}^{6}+\beta d_{2}^{6}\right),\label{eq:a23}\\
A_{31}=n_{33} m_{11}-n_{13} m_{31}=\alpha\left(d_{1}^{7}+\beta d_{2}^{7}\right),\label{eq:a31}\\
A_{32}=n_{33} m_{12}-n_{13} m_{32}=\alpha\left(d_{1}^{8}+\beta d_{2}^{8}\right),\label{eq:a32}\\
A_{33}=n_{33} m_{13}-n_{13} m_{33}=\alpha\left(d_{1}^{9}+\beta d_{2}^{9}\right),\label{eq:a33}\\
B_{11}=n_{31} m_{21}-n_{21} m_{31}=\alpha\left(d_{1}^{10}+\beta d_{2}^{10}\right),\label{eq:b11}\\
B_{12}=n_{31} m_{22}-n_{21} m_{32}=\alpha\left(d_{1}^{11}+\beta d_{2}^{11}\right),\label{eq:b12}\\
B_{13}=n_{31} m_{23}-n_{21} m_{33}=\alpha\left(d_{1}^{12}+\beta d_{2}^{12}\right),\label{eq:b13}\\
B_{21}=n_{32} m_{21}-n_{22} m_{31}=\alpha\left(d_{1}^{13}+\beta d_{2}^{13}\right),\label{eq:b21}\\
B_{22}=n_{32} m_{22}-n_{22} m_{32}=\alpha\left(d_{1}^{14}+\beta d_{2}^{14}\right),\label{eq:b22}\\
B_{23}=n_{32} m_{23}-n_{22} m_{33}=\alpha\left(d_{1}^{15}+\beta d_{2}^{15}\right),\label{eq:b23}\\
B_{31}=n_{33} m_{21}-n_{23} m_{31}=\alpha\left(d_{1}^{16}+\beta d_{2}^{16}\right),\label{eq:b31}\\
B_{32}=n_{33} m_{22}-n_{23} m_{32}=\alpha\left(d_{1}^{17}+\beta d_{2}^{17}\right),\label{eq:b32}\\
B_{33}=n_{33} m_{23}-n_{23} m_{33}=\alpha\left(d_{1}^{18}+\beta d_{2}^{18}\right),\label{eq:b33}\\
n_{31}=\alpha\left(d_{1}^{19}+\beta d_{2}^{19}\right),\label{eq:n31}\\
n_{32}=\alpha\left(d_{1}^{20}+\beta d_{2}^{20}\right),\label{eq:n32}\\
n_{33}=\alpha\left(d_{1}^{21}+\beta d_{2}^{21}\right),\label{eq:n33}\\
m_{31}=\alpha\left(d_{1}^{22}+\beta d_{2}^{22}\right),\label{eq:m31}\\
m_{32}=\alpha\left(d_{1}^{23}+\beta d_{2}^{23}\right),\label{eq:m32}\\
m_{33}=\alpha\left(d_{1}^{24}+\beta d_{2}^{24}\right).\label{eq:m33}
\end{eqnarray}
Combining~\Eref{eq:a11}, \Eref{eq:a13}, \Eref{eq:a31}, and \Eref{eq:a33}, we obtain
\begin{equation}
n_{31} A_{31} m_{33}-n_{31} A_{33} m_{31}-n_{33} A_{11} m_{33}+n_{33} A_{13} m_{31}=0.
\label{eq:combine1}
\end{equation}
Rewriting~\Eref{eq:combine1} in terms of elements of $\mathbf{d}_{1}$ and $\mathbf{d}_{2}$ gives
\begin{equation}
\begin{split}
&\alpha^{3}\left(d_{1}^{19}+\beta d_{2}^{19}\right)\left(d_{1}^{7}+\beta d_{2}^{7}\right)\left(d_{1}^{24}+\beta d_{2}^{24}\right) \\
&-\alpha^{3}\left(d_{1}^{19}+\beta d_{2}^{19}\right)\left(d_{1}^{9}+\beta d_{2}^{9}\right)\left(d_{1}^{22}+\beta d_{2}^{22}\right) \\
&-\alpha^{3}\left(d_{1}^{21}+\beta d_{2}^{21}\right)\left(d_{1}^{1}+\beta d_{2}^{1}\right)\left(d_{1}^{24}+\beta d_{2}^{24}\right) \\
&+\alpha^{3}\left(d_{1}^{21}+\beta d_{2}^{21}\right)\left(d_{1}^{3}+\beta d_{2}^{3}\right)\left(d_{1}^{22}+\beta d_{2}^{22}\right)= 0.
\label{eq:est_beta}
\end{split}
\end{equation}
Note that $\alpha$ is canceled out on both sides of~\Eref{eq:est_beta} and $\beta$ is the only variable involved in the final polynomial equation, which can be simply solved using Matlab.

Substituting the obtained $\beta$ in~\Eref{eq:w_sol}, $\mathbf{W}$ is reformulated as $\mathbf{W}=\alpha \mathbf{d}$, where $\mathbf{d}=\mathbf{d}_{1}+\beta \mathbf{d}_{2}=\left(d_{1},\ d_{2},\ \cdots,\ d_{24}\right)^{\mathrm{T}}$. The solution of $\mathbf{W}$ is in the solution space spanned by only one basis.

If $m_{31} \neq 0$, ~\Eref{eq:a11}, ~\Eref{eq:a21}, ~\Eref{eq:a31}, ~\Eref{eq:a31}, ~\Eref{eq:b11}, ~\Eref{eq:b21}, and ~\Eref{eq:b31} are reformatted as
\begin{gather}
n_{11}=\frac{-A_{11}+n_{31} m_{11}}{m_{31}}=\frac{-d_{1}+d_{19} m_{11}}{d_{22}}, 
\label{eq:n11}\\
n_{12}=\frac{-A_{21}+n_{32} m_{11}}{m_{31}}=\frac{-d_{4}+d_{23} m_{11}}{d_{22}},
\label{eq:n12}\\
n_{13}=\frac{-A_{31}+n_{33} m_{11}}{m_{31}}=\frac{-d_{7}+d_{21} m_{11}}{d_{22}},
\label{eq:n13}\\
n_{21}=\frac{-B_{11}+n_{31} m_{21}}{m_{31}}=\frac{-d_{10}+d_{19} m_{21}}{d_{22}},
\label{eq:n21}\\
n_{22}=\frac{-B_{21}+n_{32} m_{21}}{m_{31}}=\frac{-d_{13}+d_{20} m_{21}}{d_{22}},
\label{eq:n22}\\
n_{23}=\frac{-B_{31}+n_{33} m_{21}}{m_{31}}=\frac{-d_{16}+d_{21} m_{21}}{d_{22}}.
\label{eq:n23}
\end{gather}
The first two columns of rotation matrices $\mathcal{M}$, $\mathcal{N}$ satisfy
\begin{equation}
\left\{\begin{array}{l}{n_{11}^{2}+n_{21}^{2}+n_{31}^{2}=1} \\ {m_{11}^{2}+m_{21}^{2}+m_{31}^{2}=1}\end{array}\right..
\label{eq:orth}
\end{equation}
Let $a=\frac{n_{31}}{m_{31}}=\frac{d_{19}}{d_{22}}$. Substituting ~\Eref{eq:n11}, ~\Eref{eq:n21} and $a$ into~\Eref{eq:orth} gives
\begin{equation}
\left(\frac{-d_{1}+d_{19} m_{11}}{d_{22}}\right)^{2}+\left(\frac{-d_{10}+d_{19} m_{21}}{d_{22}}\right)^{2}-a^{2}\left(m_{11}^{2}+m_{21}^{2}\right)-1+a^{2}=0.
\label{eq:dma}
\end{equation}
From~\Eref{eq:dma}, we can obtain
\begin{equation}
m_{21}=\frac{\left(d_{1}^{2}-2 d_{1} d_{19} m_{11}+d_{10}^{2}-d_{22}^{2}+d_{19}^{2}\right)}{2 d_{10} d_{19}}.
\label{eq:m21}
\end{equation}
Substitution~\Eref{eq:m21} in~\Eref{eq:n21} and ~\Eref{eq:n22}, $n_{21}$ and $n_{22}$ can now be expressed in terms of $m_{11}$. 
According to the orthogonality of the first two columns of $\mathcal{N}$, we have the following constraint
\begin{equation}
n_{11} n_{12}+n_{21} n_{22}+n_{31} n_{32}=0.
\label{eq:orth_n}
\end{equation}
Since \Eref{eq:orth_n} has only two unknowns $\alpha$ and $m_{11}$, we can then represent $m_{11}$ in terms of $\alpha$. Let $m_{11}^{\alpha}$ be $m_{11}$ represented in terms of $\alpha$, which we can obtain by the Matlab command: $m_{11}^{\alpha}=\operatorname{solve}\left(n_{11} n_{12}+n_{21} n_{22}+\alpha^{2} d_{19} d_{20}, m_{11}\right)$.
We can similarly obtain $n_{11}^{\alpha}$, $n_{12}^{\alpha}$, $n_{13}^{\alpha}$, $m_{21}^{\alpha}$, $n_{21}^{\alpha}$, $n_{22}^{\alpha}$, and $n_{23}^{\alpha}$.
Since the second column of $\mathcal{N}$ satisfies $n_{12}^{2}+n_{22}^{2}+n_{32}^{2}=1$, $\alpha$ can be solved by $\alpha_{s o l}=\operatorname{solve}\left(\left(n_{12}^{\alpha}\right)^{2}+\left(n_{22}^{\alpha}\right)^{2}+\alpha^{2} d_{20}^{2}-1, \alpha\right)$.
Given $\alpha$ and $\beta$, $\mathbf{W}$ can then be obtained. The relative poses for the reference plane are then extracted. 

If $m_{31} = 0$, the above derivation can be further simplified to obtain the results similarly.